%% file: main.tex
\documentclass[acmsmall,screen,natbib=true]{acmart}

\usepackage{rotating}
\usepackage{bbm}
\usepackage{multirow}
\usepackage{tablefootnote}
\usepackage{threeparttable} 

\usepackage[linesnumbered, ruled]{algorithm2e}
\usepackage{algpseudocode}

\usepackage{natbib}

\usepackage{diagbox}
\usepackage{colortbl}
\usepackage{booktabs}
\usepackage{hhline}

\usepackage{xcolor}
\usepackage{tikz}

\definecolor{low-color}{rgb}{1,0.949,0.8}
\definecolor{mid-color}{rgb}{0.945,0.761,0.196}
\definecolor{high-color}{rgb}{0.749,0.565,0}

\newcommand{\colorboxed}[3]{\begingroup
\setlength{\fboxsep}{#1}
\colorlet{currentcolor}{.}
\def\boxcolor{#2}
\def\bgcolor{#3}
\color{#2}%
\boxed{\color{currentcolor}\colorbox{#3}{\phantom{\rule{#1}{#1}}\kern-\fboxsep\hbox{#1}\kern-\fboxsep\phantom{\rule{#1}{#1}}}}%
\endgroup}

\usepackage{arydshln}

\usepackage{pifont}
\usepackage{hyperref}
\usepackage{enumitem}
\usepackage{dashrule}
\usepackage{tikz}

\usepackage{vcell}

\usepackage{pifont} 

\usepackage{amsthm}

\usepackage{tcolorbox}
\tcbuselibrary{most}
\newtcbtheorem[]{mydef}{{Definition}}%
  {colback=yellow!20,colframe=blue!50!black,fonttitle=\bfseries}{th}

\usepackage{titletoc}
\titlecontents{section}
    [1.5em] 
    {\addvspace{0pc}\bfseries} 
    {\contentslabel{1.5em}} 
    {\hspace*{-1.5em}} 
    {\titlerule*[.5pc]{.}\contentspage}
\titlecontents{subsection}
    [4em]
    {\small\bfseries}
    {\contentslabel{2.8em}}
    {\hspace*{-2.8em}}
    {\titlerule*[.5pc]{.}\contentspage}  

\newcommand{\etal}{et al.~}
\newcommand{\eg}{e.g.,~}
\newcommand{\ie}{i.e.,~}
\newcommand{\etc}{etc.}

\newcommand{\gsb}[1]{\textcolor{teal}{#1}}
\newcommand{\gda}{GDAug}

\newcommand{\red}[1]{\textcolor{red}{#1}}

\AtBeginDocument{%
  }

\setcopyright{acmcopyright}
\copyrightyear{2018}
\acmYear{2018}
\acmDOI{XXXXXXX.XXXXXXX}

\acmJournal{JACM}
\acmVolume{37}
\acmNumber{4}
\acmArticle{111}
\acmMonth{8}

\begin{document}

\title{Data Augmentation on Graphs: A Technical Survey}

\author{Jiajun Zhou}
\authornote{Jiajun Zhou and Qi Xuan are the corresponding authors.}
\affiliation{%
  \institution{Zhejiang University of Technology}
  \city{Hang Zhou}
  \country{China}}
\email{jjzhou@zjut.edu.cn}

\author{Chenxuan Xie}
\affiliation{%
  \institution{Zhejiang University of Technology}
  \city{Hang Zhou}
  \country{China}}
\email{hello.crabboss@gmail.com}

\author{Shengbo Gong}
\affiliation{%
  \institution{Zhejiang University of Technology}
  \city{Hang Zhou}
  \country{China}}
\email{jshmhsb@gmail.com}


\author{Zhenyu Wen}
\affiliation{%
  \institution{Zhejiang University of Technology}
  \city{Hang Zhou}
  \country{China}}
\email{wenluke427@gmail.com}

\author{Xiangyu Zhao}
\affiliation{%
  \institution{City University of Hong Kong}
  \city{Hong Kong}
  \country{China}}
\email{xy.zhao@cityu.edu.hk}

\author{Qi Xuan}
\authornotemark[1]
\affiliation{%
  \institution{Zhejiang University of Technology}
  \city{Hang Zhou}
  \country{China}}
\email{xuanqi@zjut.edu.cn}

\author{Xiaoniu Yang}
\affiliation{%
  \institution{Zhejiang University of Technology}
  \city{Hang Zhou}
  \country{China}}
\email{yxn2117@126.com}
\renewcommand{\shortauthors}{Zhou et al.}

\begin{abstract}
In recent years, graph representation learning has achieved remarkable success while suffering from low-quality data problems. As a mature technology to improve data quality in computer vision, data augmentation has also attracted increasing attention in graph domain. To advance research in this emerging direction, this survey provides a comprehensive review and summary of existing graph data augmentation (\gda) techniques. Specifically, this survey first provides an overview of various feasible taxonomies and categorizes existing \gda~ studies based on multi-scale graph elements. Subsequently, for each type of \gda~ technique, this survey formalizes standardized technical definition, discuss the technical details, and provide schematic illustration. The survey also reviews domain-specific graph data augmentation techniques, including those for heterogeneous graphs, temporal graphs, spatio-temporal graphs, and hypergraphs. In addition, this survey provides a summary of available evaluation metrics and design guidelines for graph data augmentation. Lastly, it outlines the applications of \gda~ at both the data and model levels, discusses open issues in the field, and looks forward to future directions. The latest advances in \gda~ are summarized in GitHub\footnote{Github Page: \url{https://github.com/jjzhou012/GDAug-Survey}}.
\end{abstract}


\maketitle

\tableofcontents
\input{1-Introduction.tex}

\input{2-Background.tex}
\input{3-Taxonomy.tex}

\input{4-1-feature.tex}

\input{4-2-node.tex}

\input{4-3-edge.tex}

\input{4-4-subgraph.tex}

\input{4-5-graph.tex}

\input{4-6-label.tex}

\input{4-main_table.tex}
\input{5-domain.tex}

\input{6-metric.tex}

\input{7-application.tex}

\input{8-challenge-future.tex}

\input{9-conclusion.tex}

\begin{acks}
  This work was supported in part by the Key R\&D Program of Zhejiang under Grants 2022C01018 and 2024C01025, by the National Natural Science Foundation of China under Grants 62103374 and U21B2001.
\end{acks}

\bibliographystyle{ACM-Reference-Format}
\bibliography{ref, ref-preprint}

\appendix


\end{document}

%% file: 1-Introduction.tex
\section{Introduction}
Graphs or networks are important data structures extensively employed for modeling diverse complex interaction systems in real-world scenarios.
For example, user interactions on Facebook can be modeled as a social network, wherein nodes correspond to accounts and edges signify the presence of a friendship between two users~\cite{campbell2013socialnetwork};
the structure of a compound can be depicted as a molecular graph, in which nodes represent atoms and edges signify the chemical bonds connecting them~\cite{MoCL-2021,SMICLR-2022};
literature databases can be modeled as citation networks, where nodes denote authors and papers, while edges capture collaboration relationships among authors, ownership connections between literature and authors, as well as citation associations between papers~\cite{greenberg2009citationnetwork}.
To effectively analyze these relational data, graph representation learning (GRL) methods have emerged as fundamental techniques, achieving remarkable success in various downstream tasks.

From a data-driven perspective, GRL relies on an ample supply of high-quality data to effectively characterize the underlying information of graphs. However, modeling real-world interaction systems often encounters various data-level challenges that detrimentally impact the learning process of graph models and their performance on downstream tasks:
{1) \textbf{Label Scarcity}: Owing to the exorbitant cost associated with data labeling, numerous domains encounter a dearth of labeled data, rendering graph learning models susceptible to overfitting. For example, the anonymity of blockchain leads to a scarcity of account identity labels in cryptocurrency transaction networks;
2) \textbf{Data Incompleteness}: Due to privacy policies and data collection losses, real-world graph data is often incomplete, wherein the absence of nodes or edges can significantly impact the performance of graph learning models. For example, privacy protection clauses in social networks may lead to missing user relationship data;
3) \textbf{Data Noise}: Real-world graph data often exhibits a significant level of noise, which can impede the training process of graph learning models and consequently diminish prediction accuracy. For instance, sensor failures within traffic networks may lead to erroneous collection of traffic data;
4) \textbf{Data Complexity}: Graph data often encompasses diverse node and edge types, along with intricate and time-varying information, posing challenges for traditional graph learning models to effectively handle. For instance, knowledge graphs and citation networks typically exhibit multiple node and edge types; real-time traffic data in transportation networks undergoes frequent changes, necessitating the consideration of spatiotemporal dynamics;
5) \textbf{Data Distribution Shift}: The inconsistency in data distribution encountered by graph learning models during training and testing can result in a significant performance degradation. For instance, user behavior data in recommendation systems may exhibit varying patterns at different time intervals, thereby introducing unseen user behaviors during the testing phase.}

Inspired by the remarkable success of data augmentation in computer vision (CV)~\cite{DA-survey} and natural language processing (NLP)~\cite{feng2021surveynlp}, various data-level challenges in the graph domain can also be addressed by developing graph-specific data augmentation.
Data augmentation can address the limited availability of training data by introducing slight modifications to existing data or generating synthetic data, thereby aiding machine learning models in alleviating the risk of overfitting during the training phase~\cite{DA-survey}.
However, unlike image and text data, graph-structured data are non-Euclidean and discrete in nature. Their semantics and topological structure are interdependent, posing challenges for reusing existing data augmentation techniques or designing novel ones.
Despite recent advances in graph data augmentation (\gda) techniques, this emerging research field remains underdeveloped and lacks: 
1) systematic taxonomy;
2) standardized technical definitions;
3) scientific evaluation metrics;
4) specific application summarization. 
Consequently, researchers face challenges in obtaining a clear and comprehensive understanding of \gda, which hampers their ability to select or design effective \gda~ techniques.

{\textbf{Comparison with Existing Related Surveys: }
Recently, several surveys have touched upon \gda~ to varying extents, as summarized in Table~\ref{tb: contributions}.
Some of them focus specifically on the theme of \gda, reviewing existing techniques according to different taxonomies, such as graph learning tasks (node-level, edge-level, graph-level)~\cite{Survey-GDAug-1}, graph element types (structure-oriented, features-oriented, labels-oriented)~\cite{Survey-GDAug-2,Survey-GSSL-1,Survey-GSSL-2,Survey-GSSL-3,Survey-GSSL-4}, learnability (rule-based, learned)~\cite{Survey-GDAug-3}, and augmented scales (micro-level, meso-level, macro-level)~\cite{Survey-GDAug-4}.
However, these surveys do not concentrate on standardized technical definitions or discussing technical details in conjunction with definitions, nor do they effectively visualize the operational process of \gda~ techniques.
Meanwhile, another part of surveys mainly focus on graph self-supervised learning~\cite{Survey-GSSL-1,Survey-GSSL-2,Survey-GSSL-3,Survey-GSSL-4}, discussing \gda~ merely as a module within graph contrastive learning, thereby lacking a broader application and technical investigation of \gda.
Additionally, a small portion of surveys focus on data-centric graph learning, only briefly mentioning \gda~ as part of the graph learning process.
Moreover, despite the continuous emergence of \gda~ techniques, existing surveys rarely discuss evaluation metrics or design criteria for \gda~ technologies.}

In this regard, this paper comprehensively summarizes the contents related to \gda, and the main contributions can be summarized as follows:
\begin{itemize}
    \item We summarize existing taxonomies for \gda~ and review representative studies via hierarchical graph element scale taxonomy (\ie feature, node, edge, subgraph, graph and label), which facilitates researchers to understand \gda~ from various design perspectives.
    \item We generalize standardized technical definitions, discuss technical details, and provide clear schematic illustrations for each type of \gda~ method. To the best of our knowledge, this is the most exhaustive summary of \gda~ from a technical perspective.
    \item We review domain-specific data augmentation studies for complex graphs.
    \item We summarize the available evaluation metrics and design guidelines for \gda.
    \item We summarize the applications of \gda~ and discuss the open issues and future directions.
\end{itemize}

\begin{table}
    \renewcommand\arraystretch{1}
    \centering
    \caption{Comparison of contributions with related surveys. 
    \raisebox{0.5ex}{\fcolorbox{black}{low-color}{}}~: Low;
    \raisebox{0.5ex}{\fcolorbox{black}{mid-color}{}}~: Medium;
    \raisebox{0.5ex}{\fcolorbox{black}{high-color}{}}~: High.}
    \label{tb: contributions}
    \resizebox{0.6\textwidth}{!}{%
    \begin{tabular}{|c|l|l|l|l|l|l|l|l|l|l|l|l|} 
    \hline
    \diagbox{\textbf{Item}}{\textbf{Survey}} & \multicolumn{1}{c|}{\rotatebox[origin=c]{90}{Ours}} & \multicolumn{1}{c|}{\rotatebox[origin=c]{90}{\cite{Survey-GDAug-1}}}  & \multicolumn{1}{c|}{\rotatebox[origin=c]{90}{\cite{Survey-GDAug-2}}} & \multicolumn{1}{c|}{\rotatebox[origin=c]{90}{\cite{Survey-GDAug-3}}} & \multicolumn{1}{c|}{\rotatebox[origin=c]{90}{\cite{Survey-GDAug-4}}}& \multicolumn{1}{c|}{\rotatebox[origin=c]{90}{\cite{Survey-GDAug-5}}} & \multicolumn{1}{c|}{\rotatebox[origin=c]{90}{\cite{Survey-GSSL-1}}} & \multicolumn{1}{c|}{\rotatebox[origin=c]{90}{\cite{Survey-GSSL-2}}} & \multicolumn{1}{c|}{\rotatebox[origin=c]{90}{\cite{Survey-GSSL-3}}} & \multicolumn{1}{c|}{\rotatebox[origin=c]{90}{\cite{Survey-GSSL-4}}} & \multicolumn{1}{c|}{\rotatebox[origin=c]{90}{\red{\cite{survey-data-centric-1}}}}  & \multicolumn{1}{c|}{\rotatebox[origin=c]{90}{\red{\cite{survey-data-centric-2}}}}   \\ 
    \hline
    Thematic Focus                                                      & {\cellcolor[rgb]{0.749,0.565,0}}$\checkmark$        & {\cellcolor[rgb]{0.749,0.565,0}}$\checkmark$                          & {\cellcolor[rgb]{0.749,0.565,0}}$\checkmark$                         & {\cellcolor[rgb]{0.749,0.565,0}}$\checkmark$                         & {\cellcolor[rgb]{0.749,0.565,0}}$\checkmark$                        & {\cellcolor[rgb]{0.749,0.565,0}}$\checkmark$                         & {\cellcolor[rgb]{1,0.949,0.8}}$\checkmark$                          & {\cellcolor[rgb]{1,0.949,0.8}}$\checkmark$                          & {\cellcolor[rgb]{1,0.949,0.8}}$\checkmark$                          & {\cellcolor[rgb]{1,0.949,0.8}}$\checkmark$                          & {\cellcolor[rgb]{1,0.949,0.8}}$\checkmark$                                   & {\cellcolor[rgb]{1,0.949,0.8}}$\checkmark$                              \\ 
    \hline                                                                                                                                                                               
    Literature Coverage                                           & {\cellcolor[rgb]{0.749,0.565,0}}$\checkmark$        & {\cellcolor[rgb]{1,0.949,0.8}}$\checkmark$                            & {\cellcolor[rgb]{0.749,0.565,0}}$\checkmark$                         & {\cellcolor[rgb]{0.749,0.565,0}}$\checkmark$                         & {\cellcolor[rgb]{1,0.949,0.8}}$\checkmark$                          & {\cellcolor[rgb]{1,0.949,0.8}}$\checkmark$                           & {\cellcolor[rgb]{1,0.949,0.8}}$\checkmark$                          & {\cellcolor[rgb]{1,0.949,0.8}}$\checkmark$                          & {\cellcolor[rgb]{1,0.949,0.8}}$\checkmark$                          & {\cellcolor[rgb]{1,0.949,0.8}}$\checkmark$                          & {\cellcolor[rgb]{1,0.949,0.8}}$\checkmark$                             & {\cellcolor[rgb]{1,0.949,0.8}}$\checkmark$                              \\ 
    \hline                                                                                                                                                                               
    Taxonomy Summary                                                    & {\cellcolor[rgb]{0.749,0.565,0}}$\checkmark$        & {\cellcolor[rgb]{1,0.949,0.8}}$\checkmark$                            & {\cellcolor[rgb]{0.945,0.761,0.196}}$\checkmark$                     & {\cellcolor[rgb]{0.945,0.761,0.196}}$\checkmark$                     & {\cellcolor[rgb]{1,0.949,0.8}}$\checkmark$                          & {\cellcolor[rgb]{1,0.949,0.8}}$\checkmark$                           & {\cellcolor[rgb]{1,0.949,0.8}}$\checkmark$                          & {\cellcolor[rgb]{1,0.949,0.8}}$\checkmark$                          & {\cellcolor[rgb]{1,0.949,0.8}}$\checkmark$                          & {\cellcolor[rgb]{1,0.949,0.8}}$\checkmark$                          & {\cellcolor[rgb]{1,0.949,0.8}}$\checkmark$                             & {\cellcolor[rgb]{1,0.949,0.8}}$\checkmark$                              \\ 
    \hline                                                                                                                                                                             
    Definition Summary                                                  & {\cellcolor[rgb]{0.749,0.565,0}}$\checkmark$        &                                                                       & {\cellcolor[rgb]{1,0.949,0.8}}$\checkmark$                           & {\cellcolor[rgb]{1,0.949,0.8}}$\checkmark$                           &                                                                     & {\cellcolor[rgb]{1,0.949,0.8}}$\checkmark$                           & {\cellcolor[rgb]{1,0.949,0.8}}$\checkmark$                          & {\cellcolor[rgb]{1,0.949,0.8}}$\checkmark$                          & {\cellcolor[rgb]{1,0.949,0.8}}$\checkmark$                          &                                                                     &                                                                              &                                                                              \\ 
    \hline                                                                                                                                                                                           
    Technical Details                                                   & {\cellcolor[rgb]{0.749,0.565,0}}$\checkmark$        &                                                                       & {\cellcolor[rgb]{1,0.949,0.8}}$\checkmark$                           & {\cellcolor[rgb]{1,0.949,0.8}}$\checkmark$                           &                                                                     & {\cellcolor[rgb]{1,0.949,0.8}}$\checkmark$                           & {\cellcolor[rgb]{1,0.949,0.8}}$\checkmark$                          & {\cellcolor[rgb]{1,0.949,0.8}}$\checkmark$                          & {\cellcolor[rgb]{1,0.949,0.8}}$\checkmark$                          &                                                                     & {\cellcolor[rgb]{1,0.949,0.8}}$\checkmark$                                   &                                                                            \\ 
    \hline                                                                                                                                                                                           
    Schematic Illustration                                              & {\cellcolor[rgb]{0.749,0.565,0}}$\checkmark$        & {\cellcolor[rgb]{1,0.949,0.8}}$\checkmark$                            &                                                                      &                                                                      & {\cellcolor[rgb]{1,0.949,0.8}}$\checkmark$                          &                                                                      & {\cellcolor[rgb]{1,0.949,0.8}}$\checkmark$                          & {\cellcolor[rgb]{1,0.949,0.8}}$\checkmark$                          & {\cellcolor[rgb]{1,0.949,0.8}}$\checkmark$                          &                                                                     &                                                                              &                                                                              \\ 
    \hline                                                                                                                                                                          
    Performance Comparison                                              &                                                     & {\cellcolor[rgb]{1,0.949,0.8}}$\checkmark$                            &                                                                      &                                                                      & {\cellcolor[rgb]{1,0.949,0.8}}$\checkmark$                          & {\cellcolor[rgb]{1,0.949,0.8}}$\checkmark$                           &                                                                     &                                                                     &                                                                     & {\cellcolor[rgb]{1,0.949,0.8}}$\checkmark$                          &                                                                              &                                                                              \\ 
    \hline                                                                                                                                                                          
    Evaluation Metrics                                                  & {\cellcolor[rgb]{0.749,0.565,0}}$\checkmark$        & {\cellcolor[rgb]{1,0.949,0.8}}$\checkmark$                            &                                                                      &                                                                      &                                                                     &                                                                      &                                                                     &                                                                     &                                                                     &                                                                     &                                                                              &                                                                              \\ 
    \hline                                                                                                                                                                                                   
    Application Summary                                                 & {\cellcolor[rgb]{0.749,0.565,0}}$\checkmark$        &                                                                       & {\cellcolor[rgb]{0.749,0.565,0}}$\checkmark$                         & {\cellcolor[rgb]{0.749,0.565,0}}$\checkmark$                         & {\cellcolor[rgb]{0.749,0.565,0}}$\checkmark$                        &                                                                      &                                                                     &                                                                     &                                                                     &                                                                     &                                                                              &                                                                              \\ 
    \hline                                                                                                                                                                                                   
    Challenges and Outlook                                              & {\cellcolor[rgb]{0.945,0.761,0.196}}$\checkmark$    & {\cellcolor[rgb]{1,0.949,0.8}}$\checkmark$                            & {\cellcolor[rgb]{0.749,0.565,0}}$\checkmark$                         & {\cellcolor[rgb]{0.749,0.565,0}}$\checkmark$                         & {\cellcolor[rgb]{0.749,0.565,0}}$\checkmark$                        & {\cellcolor[rgb]{0.749,0.565,0}}$\checkmark$                         &                                                                     &                                                                     &                                                                     &                                                                     &                                                                              &                                                                              \\ 
    \hline                                                                                                                                                                          
    Resource Summary                                                    & {\cellcolor[rgb]{0.749,0.565,0}}$\checkmark$        & {\cellcolor[rgb]{1,0.949,0.8}}$\checkmark$                            & {\cellcolor[rgb]{0.945,0.761,0.196}}$\checkmark$                     & {\cellcolor[rgb]{0.945,0.761,0.196}}$\checkmark$                     & {\cellcolor[rgb]{1,0.949,0.8}}$\checkmark$                          & {\cellcolor[rgb]{1,0.949,0.8}}$\checkmark$                           & {\cellcolor[rgb]{1,0.949,0.8}}$\checkmark$                          & {\cellcolor[rgb]{1,0.949,0.8}}$\checkmark$                          & {\cellcolor[rgb]{1,0.949,0.8}}$\checkmark$                          &                                                                     &                                                                              & {\cellcolor[rgb]{1,0.949,0.8}}$\checkmark$                                                                             \\
    \hline                                             
    \end{tabular}}
\end{table}

%% file: 2-Background.tex
\section{Preliminaries}
\subsection{Graph}
A graph that contains all the necessary and optional graph elements can be represented as 
$G=\left(\mathcal{V}\ , \mathcal{E}\ , \boldsymbol{X}_\textit{v}\ , \boldsymbol{X}_\textit{e}\ , \mathcal{Y}_\textit{v}, Y \right)$,
where $\mathcal{V}=\{v_1, v_2, \cdots, v_{|\mathcal{V}|}\}$ and $\mathcal{E}=\{e_1, e_2, \cdots, e_{|\mathcal{E}|} \mid e=(v_\textit{i}, v_\textit{j});\  v_\textit{i}, v_\textit{j}\in \mathcal{V}\}$ are the sets of nodes and edges respectively, 
$\boldsymbol{X}_\textit{v} \in \mathbb{R}^{{|\mathcal{V}|} \times F_\textit{v}}$ and $\boldsymbol{X}_\textit{e} \in \mathbb{R}^{{|\mathcal{E}|} \times F_\textit{e}}$ are the feature matrices of nodes and edges respectively, 
$\mathcal{Y}_\textit{v}=\{(v_\textit{i}, y_\textit{i}) \mid v_\textit{i}\in \mathcal{V}\}$ is the set of node labels and $Y$ is the graph label.
For the sake of simplicity, the subscript of the feature matrix is ignored when there is no need to specify what kind of features.
The necessary structure elements $\left(\mathcal{V}, \mathcal{E}\right)$ can also be represented alternatively as adjacency matrix $\boldsymbol{A} \in \mathbb{R}^{|\mathcal{V}| \times |\mathcal{V}|}$, where $\boldsymbol{A}_\textit{ij} = \mathbbm{1}\left[(v_\textit{i}, v_\textit{j})\in \mathcal{E} \right]$ for $1 \leq i,j\leq |\mathcal{V}|$.
A diagonal matrix $\boldsymbol{D} \in \mathbb{R}^{|\mathcal{V}| \times |\mathcal{V}|}$ defines the degree distribution of $G$, and $\boldsymbol{D}_\textit{ii}=\sum_{j=0}^{|\mathcal{V}|-1} \boldsymbol{A}_\textit{ij}$.
The main notations used in this paper are listed in Table~\ref{tb: notation}.


\subsection{Data Augmentation}
Data augmentation can increase training data without collecting or labeling more data. 
Instead, it enriches the data distribution by slightly modifying existing data or synthesizing new data.
Data augmentation serves as a regularizer to help machine learning models reduce the risk of over-fitting during the training phase, and has been widely applied in CV and NLP, such as rotation, cropping, scaling, flipping, mixup, back translation, and synonym substitution.
In the graph domain, data augmentation can be regarded as a transformation function on graphs: $f: G=(\boldsymbol{A}, \boldsymbol{X}) \rightarrow \hat{G}=(\hat{\boldsymbol{A}}, \hat{\boldsymbol{X}})$, where $\hat{G}$ is the generated augmented graph.                              
However, due to the non-Euclidean data nature and the dependencies between the semantics and topology of samples, it is challenging to transfer existing data augmentation techniques into the graph domain or design effective graph augmentation techniques.
Therefore, research and investigations on graph augmentation techniques are urgently needed and valuable.

\begin{table}
    \renewcommand\arraystretch{1.2}
    \centering
    \caption{Main notations used in this paper.}
    \label{tb: notation}
    \resizebox{\textwidth}{!}{%
    \begin{tabular}{lr|lr} 
    \hline\hline
    Notation                                                                      & Description                                               & Notation                         & Description  \\ 
    \hline                                                                           
    $G\ , \hat{G}\ , g$                                                                                   & Source graph, augmented graph, subgraph                                               & $f$                    & Model                         \\
    $\mathcal{V}\ , \hat{\mathcal{V}}$                                                                    & Node set, augmented node set                                                          & $\boldsymbol{D}$       & Degree matrix                       \\
    $\mathcal{E}\ , \hat{\mathcal{E}}$                                                                    & Edge set, augmented edge set                                                          & $|\cdot|$              & The number of elements in the set              \\
    $\boldsymbol{X}\ , \boldsymbol{X}_\textit{v} \ , \boldsymbol{X}_\textit{e}\ , \hat{\boldsymbol{X}}$   & Feature matrix, node feature matrix, edge feature matrix, Augmented feature matrix    & $k$                    & Parameter of topk algorithm    \\
    $\boldsymbol{A}\ , \hat{\boldsymbol{A}}$                                                              & Adjacency matrix, augmented adjacency matrix                                          & $\mathbbm{1}$          & Location indicator matrix     \\
    $y\ ,Y\ ,\mathcal{Y}$                                                                                 & Node label, graph label, label set                                                    & $\lambda$              & Parameter of interpolation    \\  
    $\boldsymbol{h}$                                                                                      & Node representation in feature space                                                  & $\mathbb{G}$           & Augmentation space            \\                                               
    $\boldsymbol{M}, M$                                                                                   & Masking value matrix, masking value                                                   & $\epsilon$             & Threshold                     \\
    $p, P$                                                                                                & Probability, probability distribution                                                 & $\mathcal{M}$                    & Metric                        \\
    $\boldsymbol{C}$                                                                                      & Corruption function / assignment matrix                                               & $\mathcal{L}$          & Loss                          \\
    $\mathcal{D}\ ,\mathcal{D}_\textit{l}\ , \mathcal{D}_\textit{u}$                                      & Dataset, labeled dataset, unlabeled dataset                                           & $\circ$                & Hadamard product              \\
    \hline\hline
    \end{tabular}}
\end{table}



%% file: 3-Taxonomy.tex
\section{Taxonomies}\label{sec: taxonomy}
{In this section, we first provide a clear overview of feasible taxonomies for \gda~ techniques. Furthermore, we present our proposed taxonomy and discuss it in comparison with existing ones.}

\subsection{{Overview of Existing Taxonomies}}
\subsubsection{{Coarse-grained Graph Element Type}}
{\gda~ techniques can be executed on different graph elements from a coarse-grained perspective, including graph features, structures, and labels, as discussed in several surveys~\cite{Survey-GDAug-2,Survey-GSSL-1,Survey-GSSL-2,Survey-GSSL-3,Survey-GSSL-4}.
\textbf{Feature-wise augmentation} modifies, creates, or fuses graph features, such as node features, edge features, or latent features learned by models. \textbf{Structure-wise augmentation} alters or generates new graph structures, from nodes and edges to subgraphs and entire graphs. 
\textbf{Label-wise augmentation} addresses the data-hungry issue by assigning pseudo labels or creating synthetic samples from labeled data.}

\subsubsection{{Target Graph Task}}
{\gda~ techniques can be designed for various graph tasks~\cite{Survey-GDAug-1}, including \textbf{node-level} tasks by manipulating node features or structures, \textbf{edge-level} tasks by reconstructing connectivity or edge features, and \textbf{graph-level} tasks by altering graph structures or creating synthetic views.}

\subsubsection{{Learnability}}
{\gda~ techniques can be categorized based on whether they are coupled with the graph learning process~\cite{Survey-GDAug-3}, \ie, whether \gda~ can be learnable and optimized during the graph learning process. 
\textbf{Non-learnable Augmentation} does not rely on the information provided by graph learning models and generally augments graph entities in a trivial (\eg random, heuristic and rule-based) manner.
\textbf{Learnable Augmentation} is coupled with graph learning models and relies on the information provided by them (\eg model parameters, training signals) to optimize augmentors and augment graph entities. 
}

\subsubsection{{Mechanism}}\label{sec: taxonomy-mechanism}
{\gda~ techniques can be designed via different mechanisms. 
\textbf{Manipulation-based augmentation} manipulates the feature or structure in the existing graph instances to augment graph entities, similar to imposing perturbations on existing graph elements.
\textbf{Generation-based Augmentation} augments graph entities by creating brand-new graph features or structures based on existing graph information.
\textbf{Sampling-based augmentation} involves sampling elements from existing graphs to create augmented instances.
}

\subsection{{Hierarchical Graph Element Scale Taxonomy for Data Augmentation}}
{As an auxiliary technique to assist in low-quality graph learning, the design of \gda~ typically needs to consider the graph type and target task. To help researchers better understand this field and easily apply relevant studies, after reviewing above taxonomies, we will review existing \gda~ techniques according to hierarchical scale of graph elements, for the following reasons:
1) \textbf{Graph elements provide structures and information at different scales.} 
All graph operations and target tasks revolve around graph elements of varying scales. When designing \gda, it is crucial to consider these elements and their properties first. Thus taxonomy based on hierarchical graph element scale can provide a more intuitive reference;
2) \textbf{Operations on graph elements offer higher interpretability.} 
Performing specific augmentation operations on graph elements at various scales enhances interpretability, facilitating researchers in understanding the effects and mechanisms underlying \gda~ techniques, thereby enabling more effective application and improvement of these techniques;
3) \textbf{Operations on graph elements have universality for downstream tasks.} 
By augmentation operations on graph elements at different scales, general support and improvements can be provided for various downstream target tasks. Whereas, categorizing \gda~ techniques based on downstream target tasks will lead to the reclassification of certain highly generalizable \gda~ techniques;
4) \textbf{The diversity of graph elements determines the flexibility of augmentation techniques.} By understanding and manipulating graph elements at different scales, flexible and adaptable \gda~ techniques can be designed for different graph types and application scenarios.}

{In summary, we hierarchically categorize \gda~ techniques based on the scale of graph elements they operate on into feature-level, node-level, edge-level, subgraph-level, graph-level and label-level.}

%% file: 4-1-feature.tex
\begin{figure*}[htp]
	\centering
		\includegraphics[width=\textwidth]{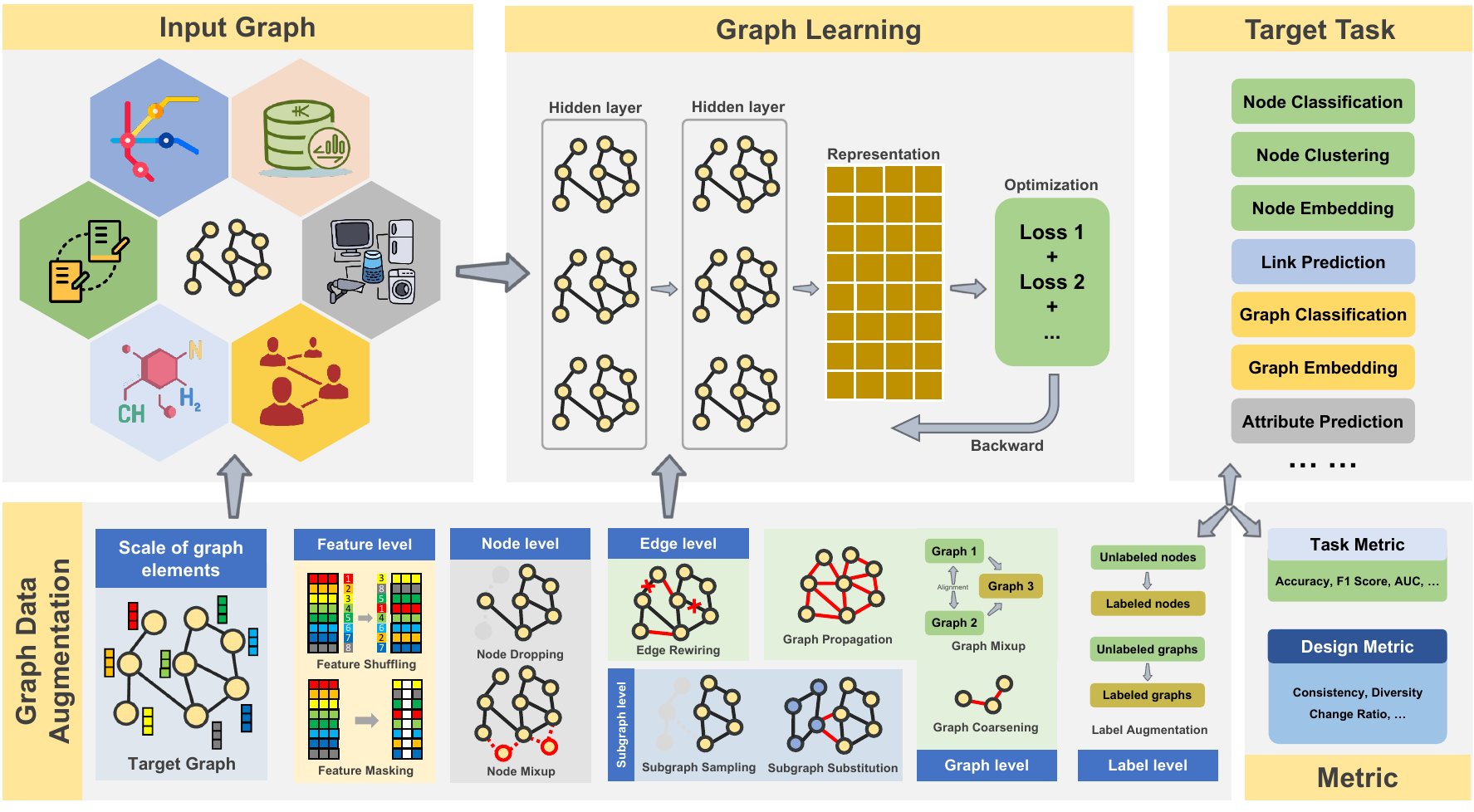}
		\caption{{An overview framework of graph data augmentation in graph representation learning.}}
		\label{fig: main}
\end{figure*}
\section{General Graph Data Augmentation}\label{sec: gda}
In this section, we review existing general \gda~ techniques for simple graphs based on our taxonomy. The overall application framework is illustrated in Fig.~\ref{fig: main}.

\subsection{Feature-level Augmentation}
The features typically consist of multiple graph attributes, which are often derived from the real physical properties of the data and play a crucial role in learning graph representations.
Graph features are available in attribute graphs and weighted graphs, and can be attached by different structure elements, such as nodes in point clouds carrying position features, edges in knowledge graphs carrying relationship information, and molecular graphs with global-level toxicological or catalytic properties.
Furthermore, embeddings or hidden features encoded by GRL models also serve as graph features.
{On this basis, feature-level \gda~ mainly involves performing various operations on the feature matrix, such as perturbing features, adding noise, \etc, while preserving the graph structure intact, so as to improve the robustness and generalization of GRL models.}

\subsubsection{\textbf{Feature Shuffling}}
{This technique aims to simulate the variation of node context information by randomly shuffling node features while preserving the graph structure, thereby creating diverse training samples and exposing models to diverse feature combinations for improved robustness and generalization.} Without loss of generality, we define feature shuffling as follows:
\begin{mydef}{~{Feature Shuffling}}{DEFexample}
    {For a given attributed graph $G=\left(\boldsymbol{A,X_\textit{v}}\right)$, feature shuffling executes row-wise shuffle on the node features, yielding an augmented graph $\hat{G}=(\boldsymbol{A}, \hat{\boldsymbol{X}}_\textit{v})$ with the same topology but rearranged nodes, \ie}
        \begin{equation}
            \hat{\boldsymbol{X}}_\textit{v} = \boldsymbol{X}_\textit{v}[\textit{idx}, :] \quad \textbf{with} \quad \textit{idx} = \textsf{Randperm}(|\mathcal{V}|)
        \end{equation}
        {where $\textsf{Randperm}(n)$ function returns a random permutation of integers from $0$ to $|\mathcal{V}|-1$.}
\end{mydef}
\begin{figure*}[!htb]
	\centering
		\includegraphics[width=0.7\textwidth]{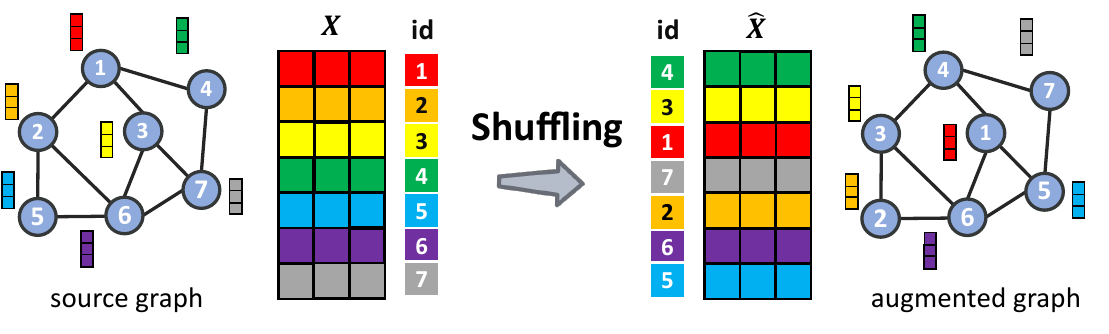}
		\caption{Illustration of feature shuffling augmentation.}
		\label{fig: feature-shuffling}
\end{figure*}
Feature shuffling augmentation is typically applied in graph contrastive learning to \textit{generate a diverse set of isomorphic negative samples}, where the topological structure is preserved but the nodes are located in different locations and receive different contextual information, as schematically depicted in Fig.~\ref{fig: feature-shuffling}. 
For example, DGI~\cite{DGI-2019} considers feature shuffling as a corruption function to generate negative samples for the first time, and plenty of related work~\cite{STABLE-2022,HDGI-2020,HDMI-2021,HTC-2023,Contrast-Reg-2023,DRGI-2023,HCHSM-2024} followed suit.
STDGI~\cite{STDGI-2019} extends the DGI method to spatio-temporal graphs and constructs negative samples by randomly permuting the node features at each time step via feature shuffling.

\subsubsection{\textbf{Feature Masking}}
{This technique aims to simulate feature uncertainty by randomly masking node features, preventing models from over-relying on specific feature patterns, thereby improving the robustness and generalization ability of models in noisy or feature-missing scenarios.}
\begin{mydef}{~{Feature Masking}}{DEFexample}
    For a given attributed graph $G=\left(\boldsymbol{A,X}\right)$, feature masking performs attribute-wise masking on graph features, 
    such that arbitrary attribute $j$ of arbitrary node $v_i$ has a probability $p_\textit{ij}^\textit{m}$ of being masked as $M_\textit{ij}$, 
    finally yielding an augmented graph $\hat{G}=(\boldsymbol{A}, \hat{\boldsymbol{X}})$ with the same topology but masked features, \ie
    \begin{equation}
        \begin{array}{c}
             \hat{\boldsymbol{X}}= \boldsymbol{X} \circ\left(\boldsymbol{1}-\mathbbm{1}_\textit{m}\right)+\boldsymbol{M} \circ \mathbbm{1}_\textit{m}  \vspace{3pt}\\
             \textbf{with} \quad \mathbbm{1}_\textit{m}[i,j] \sim \textsf{Bernoulli}\left(p_\textit{ij}^\textit{m}\right), \quad \boldsymbol{M}[i,j]=M_\textit{ij} 
        \end{array}
    \end{equation}
    where $\circ$ is the Hadamard product, $\mathbbm{1}_\textit{m}$ is the masking location indicator matrix, each element in $\mathbbm{1}_\textit{m}$ is drawn from a Bernoulli distribution with $p_\textit{ij}^\textit{m}$, and $\boldsymbol{M}$ is the masking value matrix.
\end{mydef}

\begin{table*}[htp]
    \renewcommand\arraystretch{1.3}
    \centering
    \caption{{Summary of different feature masking augmentations.}}
    \label{tb: feature-masking-strategy}
	\resizebox{\textwidth}{!}{%
    \begin{tabular}{cccc} 
    \hline\hline
    \multicolumn{1}{c}{\multirow{2}{*}{\textbf{Sub-category}}}                                                                              & \multirow{2}{*}{\textbf{Reference}}    & \multicolumn{2}{c}{\textbf{Parameter Setting}}          \\ 
    \cline{3-4}
    \multicolumn{1}{c}{}                                                                       &                                       & $p_\textit{ij}^\textit{m}$                     & $M_\textit{ij}$ / $\boldsymbol{M}$                      \\ 
    \hline            
    global uniform zero masking                               & \begin{tabular}[c]{@{}c@{}}GRACE~\cite{GRACE-2020}, BGRL~\cite{BGRL-2022}, GROC~\cite{GROC-2021}, MERIT~\cite{MERIT-2021}, \\Ethident~\cite{Ethident-2022}, GBT~\cite{GBT-2022}, CCS-SSG~\cite{CCA-SSG}, GCL-LS~\cite{GCL-LS-2024}\end{tabular}      & constant                                       & 0                                                                                   \\
    \hdashline
    local zero masking                           & HeCo~\cite{HeCo-2021}, MH-Aug~\cite{MH-Aug-2021}, GRAND~\cite{GRAND-2020}, AutoGRL~\cite{AutoGRL-2021}                                                                                                                                               & 1 or 0                                              & 0                                 \\
    \hdashline
    noise masking                                      & GraphCL~\cite{GraphCL-2020}, JOAO~\cite{JOAO-2021}, MeTA~\cite{MeTA-2021}, A2-CLM~\cite{A2-CLM-2024}                                                                                                                                                & 1                                              & $\mathcal{N}(\boldsymbol{0}, \boldsymbol{\Sigma})$  or $\mathcal{N}(\boldsymbol{X}, \boldsymbol{\Sigma})$                                 \\
    \hdashline
    \multirow{2}{*}{~~~~~importance-based masking~~~~~}   & GCA~\cite{GCA-2021}                                                                                                                                                                                                                                 & node centrality                                & 0                                                                                        \\
    \cdashline{2-4}                                                                                                                    
                                                       & NodeAug~\cite{NodeAug-2020}, InMvie~\cite{InMvie-2024}                                                                                                                                                                                               & {~~~~~attribute weight~~~~~}                   & {~~~~~more relevant attributes~~~~~}                                                      \\
    \hdashline                                                                                                                    
    gradient-based masking                             & FLAG~\cite{FLAG-2022}, A2-CLM~\cite{A2-CLM-2024}                                                                                                                                                                                                       & 1                                              & adversarial perturbation                                                              \\
    \hline\hline
    \end{tabular}}
\end{table*}

\begin{figure*}[htp]
	\centering
		\includegraphics[width=\textwidth]{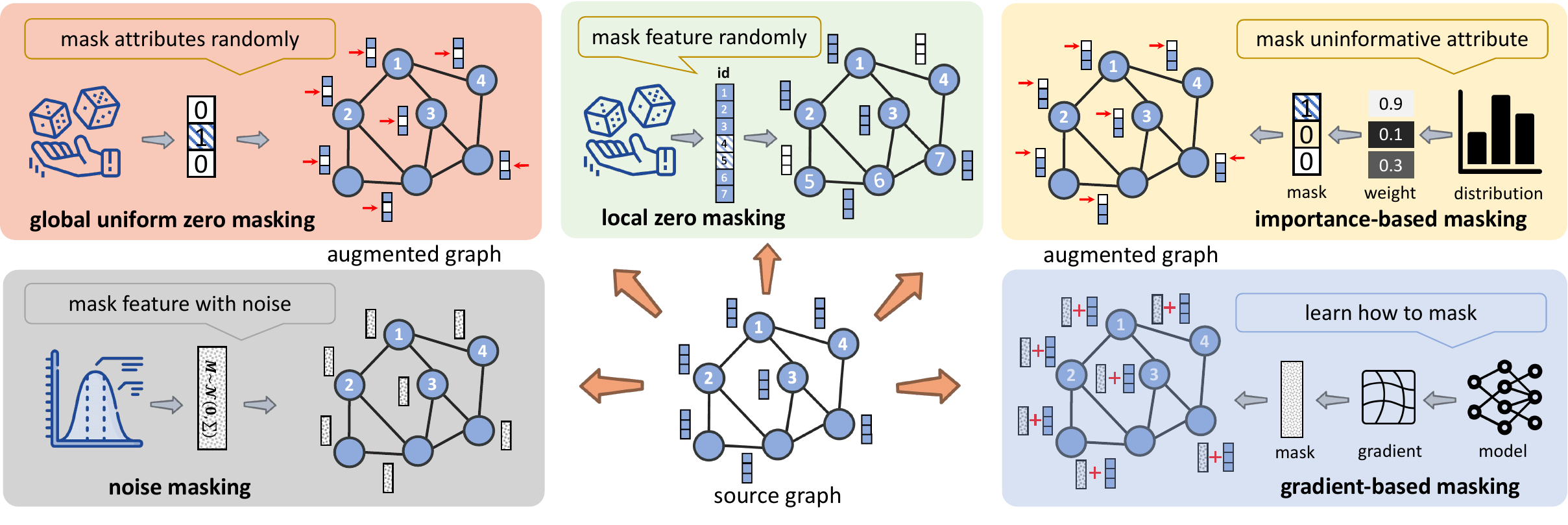}
		\caption{{Illustration of different feature masking augmentations.}}
		\label{fig: feature-masking}
\end{figure*}

Note that the elements of masking matrix $\mathbbm{1}_\textit{m}[i,j]$ are set to 1 individually with a probability $p_\textit{ij}^\textit{m}$ and 0 with a probability $1-p_\textit{ij}^\textit{m}$.
Different types of masking probabilities $p_\textit{ij}^\textit{m}$ and masking values $M_\textit{ij}$ specify different augmentation strategies, as shown in Table~\ref{tb: feature-masking-strategy} and Fig.~\ref{fig: feature-masking}.
\textbf{Global uniform zero masking} randomly masks a fraction of the dimensions with zeros in graph features, in which $p_\textit{ij}^\textit{m}$ is set to fixed constants and $M_\textit{ij}=0$, and has been extensively employed in~\cite{GRACE-2020,BGRL-2021,BGRL-2022,GROC-2021,MERIT-2021,Ethident-2022,GBT-2022} to generate contrastive views.
\textbf{Local zero masking} is another strategy that masks messages from selected nodes by setting their features to all-zero vectors. This strategy enables a node to aggregate messages only from a subset of its neighbors, achieving an augmentation in the receptive field of the target node. Consequently, it is widely applied in node-level tasks~\cite{HeCo-2021,MH-Aug-2021,GRAND-2020,AutoGRL-2021}.
\textbf{Noise masking} utilized by GraphCL~\cite{GraphCL-2020} replaces the entire node feature matrix with Gaussian noise, where $p_\textit{ij}^\textit{m}=1$ and $\boldsymbol{M}\sim \mathcal{N}(\boldsymbol{0}, \boldsymbol{\Sigma})$.
\textbf{Importance-based masking} involves selectively masking certain features of nodes in the graph based on their importance. For example, NodeAug~\cite{NodeAug-2020} replaces the uninformative attributes with more relevant ones, and computes mask probabilities through attribute weights. Similarly, GCA~\cite{GCA-2021} uses multiple node centralities to compute the mask probabilities.
\textbf{Gradient-based masking}, as presented in FLAG~\cite{FLAG-2022}, introduces learnable masking that iteratively augments node features with adversarial perturbations during training. 
Finally, it's worth noting that feature masking can be applied to various graph features like node features, edge features or other relevant features.
For example, Ethident~\cite{Ethident-2022} uniformly masks edge features in Ethereum interaction graph, MeTA~\cite{MeTA-2021} perturbs the timestamp attribute in edges, and SMICLR~\cite{SMICLR-2022} designs the `XYZ mask' augmentation to add tiny perturbations in the atoms' coordinates (\ie node position features) during molecular graph representation learning.

%% file: 4-2-node.tex
\subsection{Node-level Augmentation}
Node-level augmentation aims to enhance data diversity and improve model robustness and generalization by performing various operations on the nodes in the graph, such as node deletion, mixup, and perturbation.
It is generally applied to both node-level and graph-level tasks.

\subsubsection{\textbf{Node Dropping}} 
{The technique aims to model uncertainty in graph structure by randomly removing nodes and their associated edges, thereby facilitating the learning of effective graph representations in scenarios where structural information is deficient.}
\begin{mydef}{~{Node Dropping}}{DEFexample}
    For a given graph $G=(\boldsymbol{A}\ , \boldsymbol{X}_\textit{v})$, node dropping first selects a certain proportion of nodes $\mathcal{V}_\textit{d}$, and then discards the selected nodes and their respective connections from the graph, finally yielding an augmented graph $\hat{G}=(\hat{\boldsymbol{A}},\hat{\boldsymbol{X}})$ with the new node set $\hat{\mathcal{V}} = \mathcal{V} \setminus  \mathcal{V}_\textit{d}$, \ie
    \begin{equation}\label{eq: node-dropping-1}
        \hat{\boldsymbol{A}}=\boldsymbol{A}\left[\hat{\mathcal{V}}\ ,\hat{\mathcal{V}}\right],\quad \hat{\boldsymbol{X}}_\textit{v} = \boldsymbol{X}_\textit{v}\left[\hat{\mathcal{V}}\ ,:\right]
    \end{equation}
\end{mydef}
\begin{figure*}[htp]
	\centering
		\includegraphics[width=\textwidth]{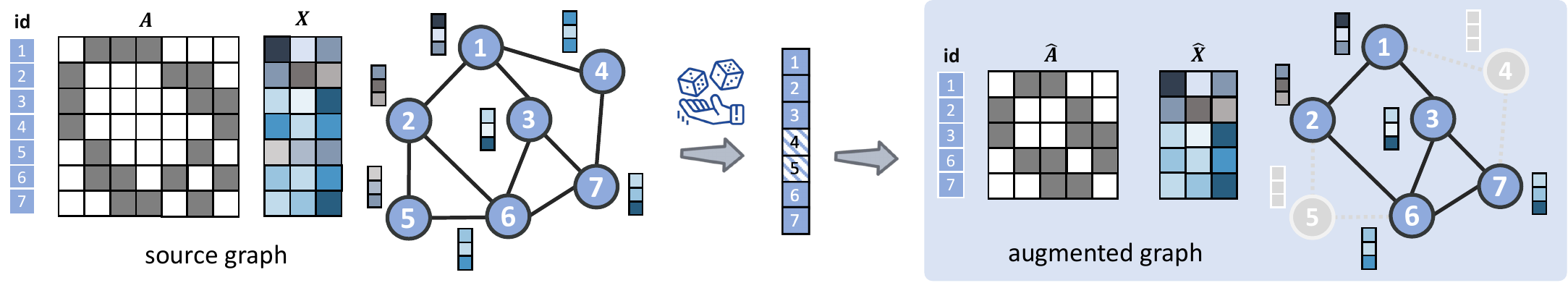}
		\caption{Illustration of node dropping augmentation. 
        }
		\label{fig: node-dropping}
\end{figure*}
Specifically, node dropping is essentially a node-level graph pruning that completely removes features and structures associated with selected nodes from the graph, eventually yielding a subgraph of the original graph as an augmented sample, as schematically depicted in Fig.~\ref{fig: node-dropping}.
This augmentation technique is generally applied in the graph classification task~\cite{GraphCL-2020,JOAO-2021,CSSL-2021,SMICLR-2022,Ethident-2022}, and can also be regarded as subgraph sampling augmentation (as discussed in Sec.~\ref{sec: subgraph-sampling}).

\subsubsection{\textbf{Node Mixup}}\label{sec: node-interpolation}
{This technique aims to introduce smoothness and continuity into the feature space through linear interpolation of node features and labels, thereby generating meaningful synthetic training samples to enhance the models' generalization ability to unseen data.}
Existing work based on node mixup is mainly inspired by techniques such as Mixup~\cite{Mixup,Manifold-Mixup} and SMOTE~\cite{SMOTE}.
Mixup is a recently proposed image augmentation technique based on the principle of Vicinal Risk Minimization (VRM), which can generate new synthetic images via linear interpolation.
Incorporating the prior knowledge that linear interpolation of features should lead to linear interpolation of the associated targets, Mixup can extend the training distribution as follows~\cite{Mixup}:
\begin{equation}
    \begin{aligned}
    &\hat{\boldsymbol{x}}=(1-\lambda) \cdot \boldsymbol{x}_\textit{i}+ \lambda \cdot \boldsymbol{x}_\textit{j} \\
    &\hat{{y}}=(1-\lambda) \cdot {y}_\textit{i} + \lambda \cdot {y}_\textit{j}
    \end{aligned}
\end{equation}
where $(\boldsymbol{x}_\textit{i}, y_\textit{i})$ and $(\boldsymbol{x}_\textit{j}, y_\textit{j})$ are two labeled samples sampled from the training set, and $\lambda \in [0,1]$.
Similarly, Manifold Mixup~\cite{Manifold-Mixup} performs mixup on the intermediate embedding space.
As for SMOTE~\cite{SMOTE}, the most popular over-sampling method, it generates new samples by performing interpolation between samples in minority classes and their nearest neighbors.
In some ways, SMOTE can be regarded as a special case of Mixup.
After reviewing their applications in the graph domain, we define node mixup augmentation as follows:
\begin{mydef}{~{Node Mixup}}{DEFexample}
    For a given attributed graph $G=(\boldsymbol{A}, \boldsymbol{X}_\textit{v}, \mathcal{Y}_\textit{v})$ and an anchor node $(v_\textit{a}, y_\textit{a})$, node mixup first samples a target node $(v_\textit{t}, y_\textit{t}) $, and then mixes the two nodes in augmentation space via linear interpolation, yielding new synthetic node $\left(\hat{v}, \hat{y}\right)$, \ie
    \begin{equation}
        \begin{array}{c}
                \hat{\boldsymbol{h}} = (1- \Lambda ) \circ \boldsymbol{h}_\textit{a} + \Lambda \circ \boldsymbol{h}_\textit{t} \vspace{3pt}\\
                \hat{y} = (1- \lambda ) \cdot y_\textit{a} + \lambda \cdot y_\textit{t} \\
        \end{array}
    \end{equation}
    where $\Lambda = \lambda \cdot (1-\mathbbm{1}_\textit{m})$, $\lambda$ is a random variable drawn from beta distribution, $\mathbbm{1}_\textit{m}$ is an optional masking location indicator vector, $\boldsymbol{h}$ is node representation in augmentation space.
\end{mydef}

\begin{figure*}[htp]
	\centering
    \includegraphics[width=\textwidth]{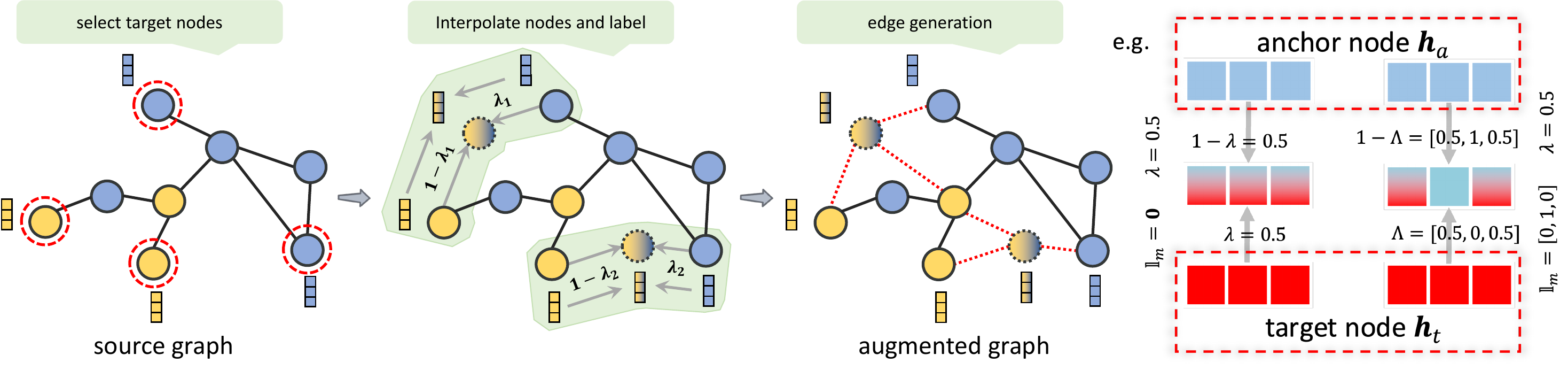}
    \caption{{Illustration of generalized node mixup augmentation.}}
    \label{fig: node-interpolation}
\end{figure*}

We show the general process of node mixup in Fig.~\ref{fig: node-interpolation}.
Note that this definition has a slightly different form from Mixup, but it is easier to incorporate existing related work, as summarized in Table~\ref{tb: node-interpolation-strategy}.
Node mixup is mainly proposed to alleviate low generalization, especially for class imbalance problem.
For example, to improve class-imbalanced node classification, existing work~\cite{GraphENS-2021,GraphMixup-2022,Graphsmote-2021} utilizes node mixup to generate synthetic nodes for minority classes. During data selection, the anchor nodes are sampled from the target minority class, and the target nodes vary for different methods. GraphMixup~\cite{GraphMixup-2022} and Graphsmote~\cite{Graphsmote-2021} consider the nearest neighbor of the anchor node with the same label as the target node.
GraphENS~\cite{GraphENS-2021} argues that selecting similar neighbors as target nodes in a highly imbalanced scenario will lead to information redundancy, so it selects target nodes from all classes.
{NodeMixup~\cite{Nodemixup-2024} addresses the under-reaching issue by selecting limited labeled nodes as anchors and sampling unlabeled nodes that are pseudo-labeled using prediction as target nodes, where the sampling probability is designed based on the neighborhood label distribution.
S-Mixup-n~\cite{S-Mixup-n-2023} performs inter-class interpolation on nodes with moderate prediction confidence from different classes, while nodes with high and low prediction confidence from the same class are used for intra-class interpolation.}
Additionally, some studies~\cite{2branch-mixup,Graphmix-2021,NodeAug-INS} do not strictly limit the selection of nodes and typically perform random sampling from training set.
\begin{table*}[htp]
    \renewcommand\arraystretch{1.4}
    \centering
    \caption{Summary of different node mixup augmentations.}
    \label{tb: node-interpolation-strategy}
    \resizebox{\textwidth}{!}{%
    \begin{tabular}{cccccc} 
    \hline\hline
    \multirow{2}{*}{\textbf{Reference}}                                  & \multicolumn{3}{c}{\textbf{Parameter Setting}}                                                                                                                                                                                                                                                                                      & \multirow{2}{*}{\begin{tabular}[c]{@{}c@{}}\textbf{Augmentation}\\\textbf{Space}\end{tabular}}                                & \multirow{2}{*}{\begin{tabular}[c]{@{}c@{}}\textbf{Adjoint Edge}\\\textbf{Generation}?\end{tabular}}               \\ 
    \cline{2-4}                    
                                                                          & Anchor Node                              & Target Node                                                                                                                                                                          & $\mathbbm{1}_\textit{m}$                             &                                &              \\ 
    \hline                                                                     
    Two-branch-Mixup~\cite{2branch-mixup}                                    & \multicolumn{2}{c}{both are randomly sampling from $\mathcal{V}$}                                                                                                                                                                                               & $\boldsymbol{0}$                                     & input / embedding        & False             \\
    Graphmix~\cite{Graphmix-2021}                                         & \multicolumn{2}{c}{within labeled / unlabeled nodes}                                                                                                                                                          & $\boldsymbol{0}$                                     & embedding                 & False             \\
    NodeAug-INS~\cite{NodeAug-INS}                                        & \multicolumn{2}{c}{both are randomly sampling from $\mathcal{V}$}                                                                                                                                                                                               & $\boldsymbol{0}$                                     & input                     & True             \\
    {S-Mixup~\cite{S-Mixup-n-2023}}                                        & \multicolumn{2}{c}{{intra-class / inter-class nodes}}                                                                                                                                                                                               & {$\boldsymbol{0}$}                                     & {input}                     & {True}             \\
    {NodeMixup~\cite{Nodemixup-2024}}                                 & {labeled nodes}                              & {unlabeled nodes}                                                                                                                                                                                    & {$\boldsymbol{0}$}                               & {input}                     & {True}             \\                                  
    GraphENS~\cite{GraphENS-2021}                                         & target minority class                                 & all classes including unlabeled nodes                                                                                                                                                          & $\mathbbm{1}_\textit{m} \in \{0,1\}^{F_\textit{v}}$  & input                     & True             \\
    GraphMixup~\cite{GraphMixup-2022}, Graphsmote~\cite{Graphsmote-2021}  & target minority class                                 & nearest neighbor of the same class   & $\boldsymbol{0}$                                     & embedding                 & True             \\
    \hline\hline
    \end{tabular}}
\end{table*}

In addition, \emph{it is worth noting that node mixup can only generate isolated synthetic nodes, so it generally relies on an edge generation module to connect the generated nodes to the graph.}
For example, Graphsmote~\cite{Graphsmote-2021} employs a weighted inner product predictor and jointly trains it via edge reconstruction task, finally yielding binary or soft edges for synthetic nodes.
Similarly, GraphMixup~\cite{GraphMixup-2022} also adopts the same edge generator but optimizes it with three self-supervised tasks: edge reconstruction, local-path prediction and global-path prediction.
Moreover, GraphENS~\cite{GraphENS-2021} first generates the adjacent node distribution for the synthetic node by mixing those of the anchor node and the target node, and then connects the synthetic node to the graph by sampling neighbors from the distribution.
Two-branch-Mixup~\cite{2branch-mixup} does not construct an edge generator to connect synthetic nodes, but indirectly uses the adjacency information of anchor nodes and target nodes to aggregate messages for synthetic nodes.
{NodeMixup~\cite{Nodemixup-2024} performs feature interpolation on intra-class nodes while also interpolating their topological information.
S-Mixup-n~\cite{S-Mixup-n-2023} utilizes edge gradient signals generated during the GNN training process to guide the selection of edges for connecting newly generated nodes.}
As an exception, GraphMix~\cite{Graphmix-2021} trains a fully connected network jointly with GNN via weight sharing, which can avoid message aggregation while performing interpolation-based regularization, thus eliminating edge generation.

Lastly, as a generalized definition, we perform feature interpolation using a composite parameter $\Lambda$, in which $\mathbbm{1}_\textit{m}$ usually serves for masking class-specific attributes of target nodes from introducing noise, as mentioned in~\cite{GraphENS-2021}.
When the mask is not used (\ie $\mathbbm{1}_\textit{m} = \boldsymbol{0}$), $\Lambda$ degenerates to $\lambda$.

%% file: 4-3-edge.tex
\subsection{Edge-level Augmentation}
{Edge-level augmentation is based on the prior assumption that modifying edges while preserving key information of the graph structure can introduce beneficial diversity. By altering some connections within the graph, it introduces structural diversity, preventing the model from overfitting to specific connection patterns.}
Existing edge-level augmentations include edge removing, edge additions and their hybrids, which we unify as \textbf{edge rewiring} augmentation.
\begin{mydef}{~{Edge Rewiring}}{DEFexample}
    For a given graph $G=\left(\boldsymbol{A,X}\right)$ without considering edge features, edge rewiring removes or adds a portion of edges in $G$, by applying masks parameterized by two probabilities $p_\textit{ij}^\textit{-}$ and $p_\textit{ij}^\textit{+}$ to the adjacency matrix $\boldsymbol{A}$, finally yields an augmented graph $\hat{G}=(\hat{\boldsymbol{A}},\boldsymbol{X})$, \ie
    \begin{equation}
        \begin{array}{c}
            \hat{\boldsymbol{A}} = \underbrace{\boldsymbol{A} \circ \left(\boldsymbol{1} - \mathbbm{1}_\textit{r} \right)}_\text{edge removing} + \underbrace{\left(\boldsymbol{1} - \boldsymbol{A}\right) \circ \mathbbm{1}_\textit{r}}_\text{edge addition} \vspace{3pt}\\
            \textbf{with} \quad  \mathbbm{1}_\textit{r}[i,j] \sim 
            \begin{cases}
                \textsf{Bernoulli}\left(p_\textit{ij}^\textit{-}\right) & \text{if} \quad \boldsymbol{A}_\textit{ij}=1 \vspace{1ex}\\
                \textsf{Bernoulli}\left(p_\textit{ij}^\textit{+}\right) & \text{if} \quad \boldsymbol{A}_\textit{ij}=0
            \end{cases}
        \end{array}
    \end{equation}
    where $\mathbbm{1}_\textit{r}$ is the rewiring location indicator matrix, $p_\textit{ij}^\textit{-}$ represents the probability of removing edge $e_\textit{ij}$ and $p_\textit{ij}^\textit{+}$ represents the probability of connecting nodes $v_\textit{i}$ and $v_\textit{j}$.
\end{mydef}

\begin{table*}[htp]
    \renewcommand\arraystretch{1.3}
    \centering
    \caption{Summary of different edge rewiring augmentations.}
    \label{tb: edge-rewiring-strategy}
    \resizebox{\textwidth}{!}{%
    \begin{tabular}{ccc} 
    \hline\hline
    \textbf{Sub-category} & \textbf{Reference} & \textbf{Parameter Setting} ($p_\textit{ij}^\textit{-}$ and $p_\textit{ij}^\textit{+}$)                                                                             \\ 
    \hline
    uniform rewiring                    & \begin{tabular}[c]{@{}c@{}}GraphCL~\cite{GraphCL-2020}, JOAO~\cite{JOAO-2021}, CSSL~\cite{CSSL-2021}, SMICLR~\cite{SMICLR-2022}, Ethident\cite{Ethident-2022},\\BGRL~\cite{BGRL-2022}, DGI~\cite{DGI-2019}, GRACE~\cite{GRACE-2020}, GBT~\cite{GBT-2022}, MERIT~\cite{MERIT-2021}\end{tabular}      & constant                                                                   \\
    \hdashline
    importance rewiring                 & GCA~\cite{GCA-2021}, NodeAug~\cite{NodeAug-2020}, MEvolve~\cite{MEvolve-cikm,MEvolve-tnse}, RobustECD~\cite{RobustECD-2021}, Fairdrop~\cite{Fairdrop-2021}                                                                                                                                         & weighted by feature and structure information                              \\
    \hdashline
    prediction rewiring                 & STABLE~\cite{STABLE-2022}, GDC~\cite{GDC-2019}, GAUG~\cite{GAUG-2021}, AutoGRL~\cite{AutoGRL-2021}                                                                                                                                                                                                 & generated by edge predictor      \\
    \hdashline
    \multirow{2}{*}{learnable rewiring} & GROC~\cite{GROC-2021}, AD-GCL~\cite{AD-GCL-2021}                                                                                                                                                                                                                                                   & binarized by gradient information                                          \\
    \cdashline{2-3}
    & MH-Aug~\cite{MH-Aug-2021}, NeuralSparse~\cite{NeuralSparse-2020}                                                                                                                                                                                                                                                                             & sampled from a learnable distribution                                      \\
    \hline\hline
    \end{tabular}}
\end{table*}
\begin{figure*}[htp]
	\centering
		\includegraphics[width=\textwidth]{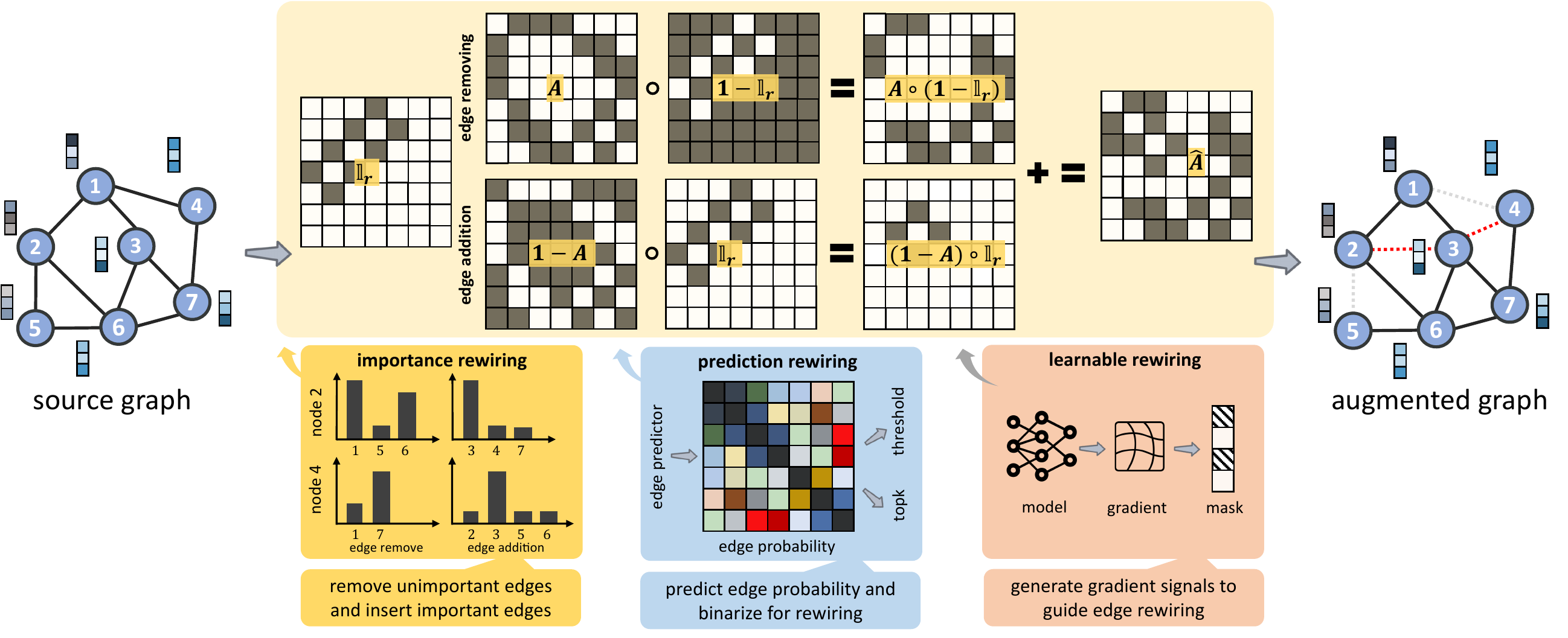}
		\caption{{Illustration of different edge rewiring augmentations.}}
		\label{fig: edge-rewiring}
\end{figure*}

Note that elements in $\mathbbm{1}_\textit{r}$ with edge locations ($\boldsymbol{A}_\textit{ij}=1$) are set to 1 individually with probability $p_\textit{ij}^\textit{-}$ and 0 otherwise, while elements in $\mathbbm{1}_\textit{r}$ with non-edge locations ($\boldsymbol{A}_\textit{ij}=0$) are set to 1 individually with probability $p_\textit{ij}^\textit{+}$ and 0 otherwise.
The first term implies edge removing where $\circ \left(\boldsymbol{1} - \mathbbm{1}_\textit{r} \right)$ drops edges $e_\textit{ij}$ if $\mathbbm{1}_\textit{r}[i,j]=1$, and the second term performs edge addition where $\boldsymbol{1} - \boldsymbol{A}$ represents the non-edge location indicator matrix and $\circ \mathbbm{1}_\textit{r}$ links node pair $(v_\textit{i}, v_\textit{j})$ if $\mathbbm{1}_\textit{r}[i,j]=1$.
In addition, when the removing probability $p_\textit{ij}^\textit{-}$ (or addition probability $p_\textit{ij}^\textit{+}$) equals to 0 for all $v_\textit{i}, v_\textit{j} \in \mathcal{V}$, the edge rewiring degrades into edge addition (or edge removing).

Furthermore, different types of rewiring probabilities specify different rewiring strategies, as summarized in Table~\ref{tb: edge-rewiring-strategy} and Fig.~\ref{fig: edge-rewiring}.
A large number of studies~\cite{DGI-2019,GRACE-2020,GBT-2022,GraphCL-2020,JOAO-2021,CSSL-2021,SMICLR-2022,Ethident-2022,BGRL-2022,MERIT-2021} related to graph contrastive learning typically use \textbf{uniform edge rewiring} to generate contrastive graph views, in which the rewiring probabilities are generally set as fixed constants.
Rong \etal~\cite{DropEdge-2020} proposed DropEdge and its layer-wise version to alleviate over-smoothing in node representation learning. The former generates a perturbed adjacency matrix via random edge removing, and shares it with all layers in the GNN models. The latter independently generates a perturbed adjacency matrix for each layer.
Similar to feature masking, rewiring strategies can also be designed in an \textbf{importance-based manner}.
The rewiring probabilities can be weighted according to different information such as node centrality~\cite{GCA-2021,NodeAug-2020}, node similarity~\cite{MEvolve-cikm,MEvolve-tnse,RobustECD-2021}, hop count~\cite{NodeAug-2020}, sensitive attribute~\cite{Fairdrop-2021}, \etc~For example, NodeAug~\cite{NodeAug-2020} considers that nodes with larger degree values and closer distance to the target node contain more information, and further weights the rewiring probabilities by node degree and hop count.
In addition, several studies first compute the edge probability matrix via different \textbf{edge prediction} strategies like node similarity~\cite{STABLE-2022}, graph diffusion~\cite{GDC-2019} and graph auto-encoder (GAE)~\cite{GAUG-2021,AutoGRL-2021}, and further achieve edge rewiring via threshold-$\epsilon$~\cite{GDC-2019} or top-$k$~\cite{STABLE-2022,GDC-2019,GAUG-2021,AutoGRL-2021}.
For example, GAUG~\cite{GAUG-2021} considers GAE as the edge predictor to generate the edge probability matrix, and then removes the top-$k$ existing edges with the least edge probabilities and adds the top-$k$ non-edges with the largest edge probabilities.

The above studies are generally regarded as non-learnable methods, and this augmentation can also be coupled with graph learning models, yielding \textbf{learnable edge rewiring}.
GROC~\cite{GROC-2021} uses gradient information to guide edge rewiring, yielding an adversarial transformation that removes a portion of edges with minimal gradient values and adds a portion of edges with maximal gradient values during model training.
ADGCL~\cite{AD-GCL-2021} designs learnable edge removing augmentation by building a random graph model, in which each edge will be associated with a random mask variable drawn from a parametric Bernoulli distribution.
MH-Aug~\cite{MH-Aug-2021} utilizes the \textit{Metropolis-Hastings} algorithm to sample the parameters of edge perturbation from target distribution, and then generates accepted augmented graphs via a designed acceptance ratio.
NeuralSparse~\cite{NeuralSparse-2020} trains a sparsification network to sample no more than $k$ edges for each node from a learned distribution, yielding a sparsified $k$-neighbors subgraph that preserves task-relevant edges.

Lastly, edge removing is more commonly used than edge addition in practice. 
The former can be regarded as a process of graph sparsification, which can 1) denoise or prune graphs by removing misleading or uninformative links; 
2) enable multiple views for random subset aggregation. 
The latter can restore missing links to a certain extent, but may also introduce noisy connections.
In addition, when the target graph contains edge weights or attributes, edge addition augmentation needs to account for the generation of additional edge information inevitably~\cite{MeTA-2021}.

%% file: 4-4-subgraph.tex
\subsection{Subgraph-level Augmentation}

{Subgraph-level augmentation is based on the prior assumption that key subgraphs can retain the overall properties of the source graph. By sampling, modifying, or combining substructures from the source graph, it creates diverse and meaningful training samples, thereby enhancing the models' ability to learn from different subgraph patterns.}

\subsubsection{\textbf{Subgraph Sampling}}\label{sec: subgraph-sampling}

{Subgraph sampling is based on the prior assumption that sampling smaller subgraphs from the source graph can preserve essential structural and feature information while reducing computational complexity. It helps models learn effectively from large-scale graphs by sampling diverse and easy-to-process subgraph samples for training.} Analogous to image cropping, subgraph sampling can be regarded as performing cropping on a graph.
Here we unify existing methods, such as subgraph sampling and graph cropping, and present a generalized definition of subgraph sampling augmentation as follows:
\begin{mydef}{~{Subgraph Sampling}}{DEFexample}
    For a given graph $G=(\boldsymbol{A}, \boldsymbol{X}_\textit{v}, \boldsymbol{X}_\textit{e})$, subgraph sampling extracts node subset $\hat{\mathcal{V}} \subseteq \mathcal{V}$ and edge subset $\hat{\mathcal{E}} \subset \mathcal{E}$ from $G$ to derive an augmented subgraph $g=\left(\hat{\mathcal{V}}, \hat{\mathcal{E}}, \hat{\boldsymbol{X}}_\textit{v}, \hat{\boldsymbol{X}}_\textit{e}\right)$, where
    \begin{equation}
             \hat{\boldsymbol{X}}_\textit{v}, \   \hat{\boldsymbol{X}}_\textit{e} =  \boldsymbol{X}_\textit{v}\left[\hat{\mathcal{V}},:\right],  \  \boldsymbol{X}_\textit{e}\left[\hat{\mathcal{E}},:\right]
    \end{equation}
\end{mydef}
\begin{table*}[htp]
    \renewcommand\arraystretch{1.3}
    \centering
    \caption{Summary of different subgraph sampling augmentations.}
    \label{tb: subgraph-sampling-strategy}
    \resizebox{\textwidth}{!}{%
    \begin{tabular}{cccc}
    \hline\hline
    \multirow{2}{*}{\textbf{Reference}}    & \multirow{2}{*}{\textbf{Sub-category}} & \multicolumn{2}{c}{\textbf{Parameter Setting}}                                                                             \\
    \cline{3-4}
                                  &                                & how to get $\hat{\mathcal{V}}$                                                & how to get $\hat{\mathcal{E}}$ and $g$         \\
    \hline
    MVGRL~\cite{MVGRL-2020}, MERIT~\cite{MERIT-2021}                               & uniform sampling         & nodes randomly sampled from a graph& $\boldsymbol{A}[\hat{\mathcal{V}},\hat{\mathcal{V}}]$    \\
    DGI~\cite{DGI-2019}, InfoGraph~\cite{InfoGraph-2020}, EGI~\cite{EGI-2021}, MEvolve~\cite{zhou2021subgraph}   & ego-net sampling         & $\{v_\textit{i}\mid \textsf{ShortestPath}(v_\textit{j}, v_\textit{i})\le l\}$  & $\boldsymbol{A}[\hat{\mathcal{V}},\hat{\mathcal{V}}]$    \\
    SUGAR~\cite{SUGAR-2021}, GraphCL~\cite{GraphCL-2020}                             & search sampling          & nodes visited by BFS/DFS starting from a node               & $\boldsymbol{A}[\hat{\mathcal{V}},\hat{\mathcal{V}}]$    \\
    GraphCL~\cite{GraphCL-2020}, JOAO~\cite{JOAO-2021}, SMICLR~\cite{SMICLR-2022}, GCC~\cite{GCC-2020}         & random-walk sampling     & nodes visited by a random walk starting from a node         & $\boldsymbol{A}[\hat{\mathcal{V}},\hat{\mathcal{V}}]$    \\
    Ethident~\cite{Ethident-2022}, SUBG-CON~\cite{SUBG-CON}, TEAug~\cite{TEAug-2022}   & importance sampling      & select top-$k$ most importance neighbors for a node& $\boldsymbol{A}[\hat{\mathcal{V}},\hat{\mathcal{V}}]$    \\
    GREA~\cite{GREA-2022}                                           & learnable sampling       & nodes selected by a learned node mask vector                & $\boldsymbol{A}[\hat{\mathcal{V}},\hat{\mathcal{V}}]$    \\  
    \hline\hline    
    \end{tabular}}  
\end{table*}
\begin{figure*}[htp]
	\centering
		\includegraphics[width=\textwidth]{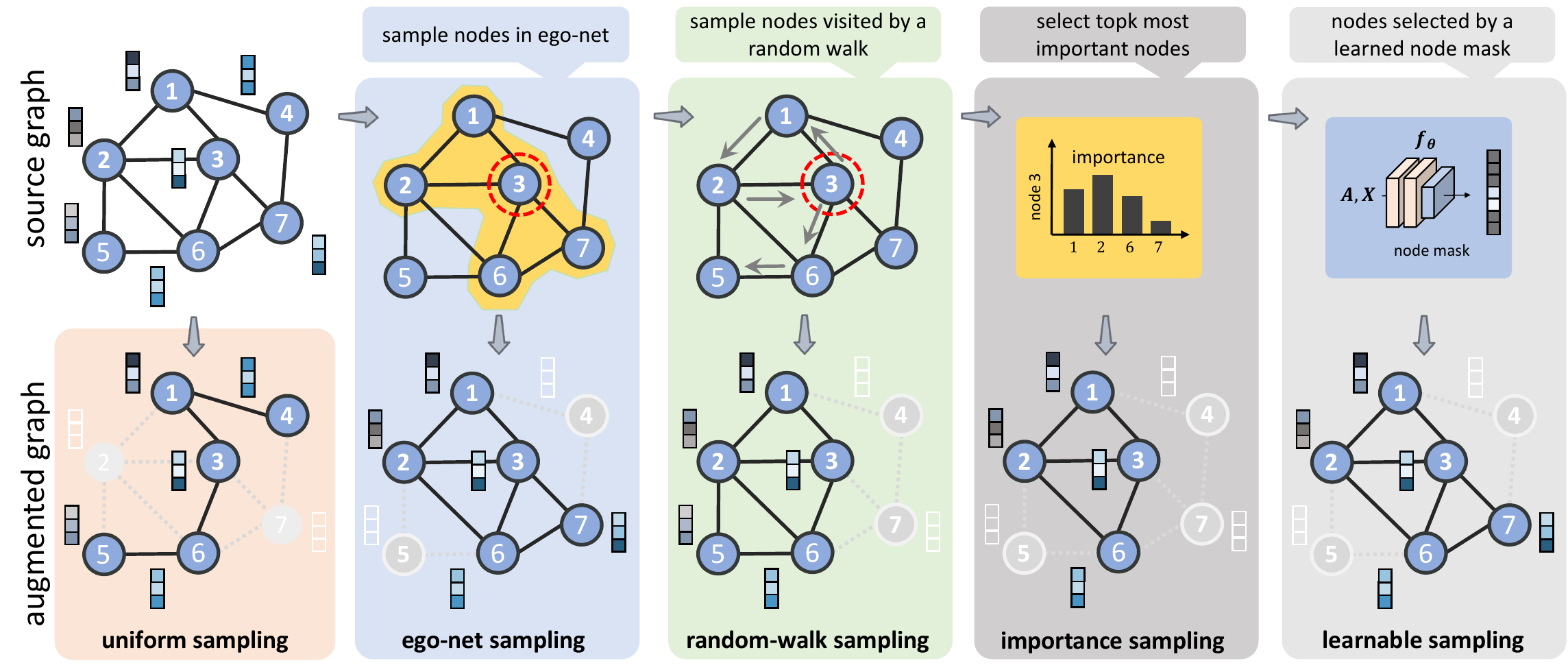}
		\caption{Illustration of different subgraph sampling augmentations.}
		\label{fig: subgraph-sampling}
\end{figure*}
In general, \emph{the process of subgraph sampling involves first selecting a subset of nodes from the graph, followed by determining the corresponding edge subset.}
It can be guided by different sampling strategies, as summarized in Table~\ref{tb: subgraph-sampling-strategy} and illustrated in Fig.~\ref{fig: subgraph-sampling}.

\textbf{Uniform sampling}~\cite{MVGRL-2020,MERIT-2021} uniformly selects a portion of nodes $\hat{\mathcal{V}}$ from $\mathcal{V}$ and then induces the subgraph topology by $\hat{\boldsymbol{A}}=\boldsymbol{A}[\hat{\mathcal{V}},\hat{\mathcal{V}}]$, which is similar to node dropping defined in Eq.~(\ref{eq: node-dropping-1}).

\textbf{Ego-net sampling} is based on the strong correlation between central nodes and their local neighborhood, and is generally used to provide patch (local) views during contrastive learning, such as in DGI~\cite{DGI-2019} and InfoGraph~\cite{InfoGraph-2020}.  
Given a node-level encoder with $l$ layers, the computation of patch representation for node $v_\textit{i}$ only depends on its $l$-hop neighborhood, aka $l$-hop ego-net.
A $l$-hop ego-net sampling for a central node $v_\textit{i}$ is essentially to obtain its receptive field subgraph with node set $\hat{\mathcal{V}} = \{v_\textit{j}\mid \textsf{ShortestPath}(v_\textit{j}, v_\textit{i})\le l\}$.
Ego-net sampling can be regarded as a special version of \textbf{Breadth-first search (BFS) sampling}, which can also be used for subgraph augmentation.
For example, SUGAR~\cite{SUGAR-2021} first selects the top-$k$ most important nodes according to degree ranking, and then extracts a subgraph for each selected node by BFS sampling.
GraphCL~\cite{GraphCL-2020} compares the performance of contrastive learning with three kinds of subgraphs, which are extracted via BFS sampling, \textbf{Depth-first search (DFS) sampling} and random-walk sampling, respectively.
It concludes that the subgraphs extracted by DFS sampling preserve less structure information but help contrastive learning achieve better performance.

\textbf{Random-walk sampling}~\cite{GraphCL-2020,JOAO-2021,SMICLR-2022,GCC-2020} is a popular strategy for extracting informative subgraphs.
A random-walk sampling starting from a given node iteratively collects node subset $\hat{\mathcal{V}}$. 
At each iteration, the walk travels to its neighborhood with the probability proportional to the edge weight.
For the random walk with restart (RWR) applied in GCC~\cite{GCC-2020}, the walk has a probability of returning to the starting node.

\textbf{Importance sampling}~\cite{Ethident-2022,SUBG-CON,Graphcrop-arxiv,NeuralSparse-2020} is proposed to extract contextual subgraphs with more structure information and less noise for given nodes.
For a given node, importance sampling first measures the importance scores of its neighbor nodes by several importance metrics or graph information, and then chooses the top-$k$ important neighbors to construct a subgraph.
For example, SUBG-CON~\cite{SUBG-CON} utilizes the Personalized PageRank centrality~\cite{pagerank-1999} to measure the node importance, while Ethident~\cite{Ethident-2022} evaluates the importance of neighbors according to different edge attributes.

In addition to the above model-agnostic augmentations, several studies have proposed \textbf{learnable sampling} strategies to automatically extract task-relevant subgraphs.
For example, GREA~\cite{GREA-2022} trains a separator that maps the node representations to a mask vector, and then uses the learned node mask to extract the rationale subgraph.


\subsubsection{\textbf{Subgraph Substitution}}
{This technique is based on the prior assumption that replacing parts of a graph with other structurally similar or functionally equivalent subgraphs can generate new graph instances while preserving the key properties of the source graph. The introduction of structural diversity enriches the training data, enabling the models to encounter a broader spectrum of graph patterns and thereby preventing overfitting.}
\begin{mydef}{~{Subgraph Substitution}}{DEFexample}
    For a pair of graphs $G=(\mathcal{V}, \mathcal{E}, Y) $ and $G^\textit{t}=(\mathcal{V}^\textit{t}, \mathcal{E}^\textit{t}, Y^\textit{t}) $, subgraph substitution first drops a subgraph $g=(\mathcal{V}_\textit{g}, \mathcal{E}_\textit{g}) $ from $G$, and then merges the remaining part with another subgraph $g^\textit{t}=(\mathcal{V}^\textit{t}_\textit{g}, \mathcal{E}^\textit{t}_\textit{g}) $ sampled from $G^\textit{t}$, finally yielding an augmented graph $\hat{G} = (\hat{\mathcal{V}}, \hat{\mathcal{E}}, \hat{Y}) $, \ie
    \begin{equation}
        \begin{array}{c}
             \hat{\mathcal{V}} = \left(\mathcal{V} \setminus \mathcal{V}_\textit{g}\right)  \cup \mathcal{V}^\textit{t}_\textit{g},\ \  \hat{\mathcal{E}} = \left(\mathcal{E} \setminus \mathcal{E}_\textit{g} \setminus \mathcal{E}^\textit{-}\right)     \cup \mathcal{E}^\textit{t}_\textit{g} \cup \mathcal{E}^\textit{+}, \ \  \hat{Y} = \left(1-\lambda\right) \cdot Y + \lambda \cdot Y^\textit{t} \vspace{1ex}\\
             \mathcal{E}^\textit{-}  = \left\{\left(v_\textit{i}, v_\textit{j}\right)  \mid \left(v_\textit{i}, v_\textit{j}\right) \in \mathcal{E}  \land  \lnot \left(v_\textit{i}, v_\textit{j} \in \mathcal{V}_\textit{g}  \right) \land   \lnot \left(v_\textit{i},v_\textit{j}\in \mathcal{V} \setminus \mathcal{V}_\textit{g} \right) \right\}   
        \end{array}
    \end{equation}
    where $\mathcal{E}^\textit{-}$ is the set of edges that break when dropping $g$ from $G$, $\mathcal{E}^\textit{+}$ is the set of edges that connect subgraph $g^\textit{t}$, and $\lambda$ is the adaptive label interpolation ratio.
\end{mydef}

\begin{table*}[htp]
    \renewcommand\arraystretch{1.3}
    \centering
    \caption{Summary of different subgraph substitution strategies.}
    \label{tb: subgraph-substitution-strategy}
    \resizebox{\textwidth}{!}{%
    \begin{tabular}{ccccc} 
    \hline\hline
    \multirow{3}{*}{\textbf{Reference}} & \multicolumn{4}{c}{\textbf{Parameter Setting}}                                                                                                                                                                              \\ 
    \cline{2-5}
                               & \multicolumn{2}{c}{\textbf{Subgraph Sampling}}                                         & \multirow{2}{*}{~~~~~~\textbf{Subgraph Merging} $(\mathcal{E}^\textit{+}, \mathcal{E}^\textit{-})$~~~~~~}                                                                      & \multirow{2}{*}{\textbf{Label Generation} ($\lambda$)}      \\ 
    \cline{2-3}
                               & $g$                                                   & $g^\textit{t}$          &                                                                                                       &                                        \\ 
    \hline
    MoCL~\cite{MoCL-2021}                          & valid molecular substructure      & bioisostere                       & $\mathcal{E}^\textit{+} = \mathcal{E}^\textit{-}$                                           & $\lambda=0$                               \\
    SubMix~\cite{SubMix-2022}                      & connected and clustered subgraph  & connected and clustered subgraph  & $\mathcal{E}^\textit{+} = \mathcal{E}^\textit{-}$                                           & $\lambda=|\mathcal{E}^\textit{t}_\textit{g}| / |\hat{\mathcal{E}}|$  \\
    GREA~\cite{GREA-2022}                          & environment subgraph              & environment subgraph              & $\mathcal{E}^\textit{+} =\emptyset$                                                         & $\lambda=0$                              \\
    \red{IGM~\cite{IGM-2024}}                            & environment subgraph              & environment subgraph              & degree-guide random edge addition                                                           & $\lambda=0$    \\
    Graph Transplant~\cite{Graph-Transplant-2022}  & partial $l$-hop ego-net           & partial $l$-hop ego net           & edge sampling or prediction              & subgraph saliency                      \\
    \hline\hline
    \end{tabular}}
\end{table*}
\begin{figure*}[htp]
	\centering
		\includegraphics[width=\textwidth]{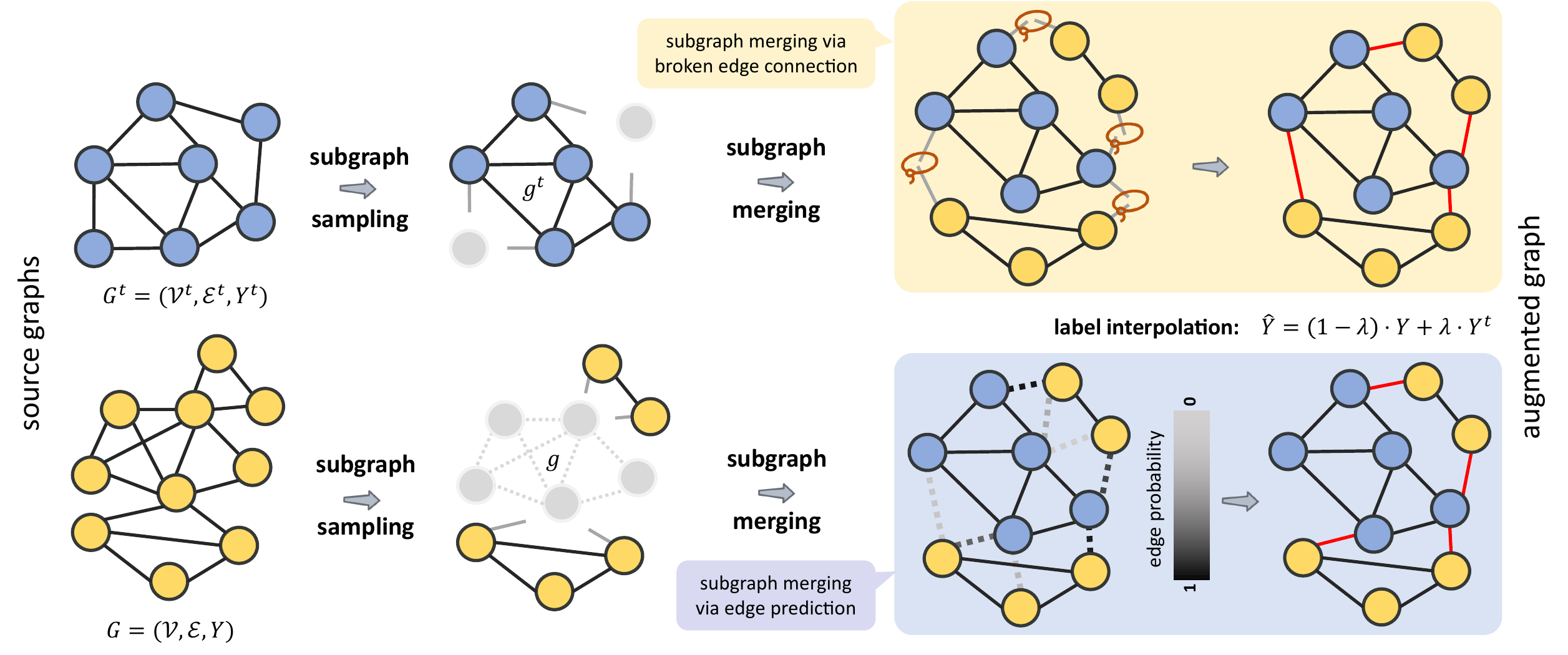}
		\caption{Illustration of generalized subgraph substitution augmentation.}
		\label{fig: subgraph-substitution}
\end{figure*}

Note that subgraph substitution is a hybrid augmentation that incorporates multiple operations like subgraph sampling, subgraph merging, and label interpolation to generate new graphs, as summarized in Table~\ref{tb: subgraph-substitution-strategy} and illustrated in Fig.~\ref{fig: subgraph-substitution}.
\emph{When performing subgraph sampling on graph pair $(G, G^\textit{t})$, the two extracted substructures $(g, g^\textit{t})$ used for substitution usually play similar roles in their respective original graphs.}
For example, MoCL~\cite{MoCL-2021} replaces functional substructures in molecules with bioisosteres~\cite{bioisostere} that share similar chemical properties.
SubMix~\cite{SubMix-2022} uses importance sampling (as described in Sec.~\ref{sec: subgraph-sampling}) to extract connected and clustered subgraphs from the given graph pair.
GREA~\cite{GREA-2022} and IGM~\cite{IGM-2024} replace the environment subgraph that can be regarded as natural noises in $G$ with another environment subgraph sampled from $G^\textit{t}$.

Furthermore, \emph{it is also essential to guarantee the connectivity of augmented graphs.}
SubMix~\cite{SubMix-2022} inserts a subgraph $g^\textit{t}$ of the same size as $g$ into $G$ without breaking the edges that connect $\mathcal{V}_\textit{g}$ with $\mathcal{V} \setminus \mathcal{V}_\textit{g}$, having $\mathcal{E}^\textit{+} = \mathcal{E}^\textit{-}$.
Similarly, MoCL~\cite{MoCL-2021} retains the chemical bonds (edges in molecular graphs) connected to $g$ before augmentation, and uses them to connect $g^\textit{t}$ during subgraph substitution.
Graph Transplant~\cite{Graph-Transplant-2022} proposes two strategies for merging subgraphs. One is uniform edge sampling to connect nodes whose degree values change during augmentation, and the other is differentiable edge prediction that considers the feature similarity of node pairs for connectivity. 
{IGM~\cite{IGM-2024} merges two subgraphs by randomly adding edges between them according to the node degree.}
As an exception, GREA~\cite{GREA-2022} performs subgraph substitution by swapping the node representations of subgraphs in the embedding space, which is free from edge generation ($\mathcal{E}^\textit{+}=\emptyset$).

Lastly, \textit{subgraph substitution generally serves for graph-level tasks, especially for graph classification, so it is necessary to assign appropriate labels for augmented graphs.}
Since $g$ and $g^\textit{t}$ play similar roles in their respective graphs, the adaptive label interpolation ratio $\lambda$ can be derived according to the contribution of $g^\textit{t}$ to the augmented graph $\hat{G}$.
For example, SubMix~\cite{SubMix-2022} assumes that edges are crucial factors in determining graph labels, and hence defines $\lambda$ as the ratio of contained edges in $g^\textit{t}$, \ie $\lambda=|\mathcal{E}^\textit{t}_\textit{g}| / |\hat{\mathcal{E}}|$.
Graph Transplant~\cite{Graph-Transplant-2022} quantifies the contribution of $g^\textit{t}$ using the total saliency of contained nodes.
In addition, MoCL~\cite{MoCL-2021}, GREA~\cite{GREA-2022} and IGM~\cite{IGM-2024} replace non-critical structures in the original graphs, so the augmented graph $\hat{G}$ and the original graph $G$ are considered to have consistent labels, \ie $g^\textit{t}$ is now a non-critical subgraph and assigned a low label weight ($\lambda = 0$).

%% file: 4-5-graph.tex
\subsection{Graph-level Augmentation}
Graph-level augmentation primarily involves transforming the entire graph to create diverse and meaningful training samples, enabling the model to better learn the global information of the graph.

\subsubsection{\textbf{Graph Propagation}}
\begin{figure*}[htp]
	\centering
		\includegraphics[width=0.9\textwidth]{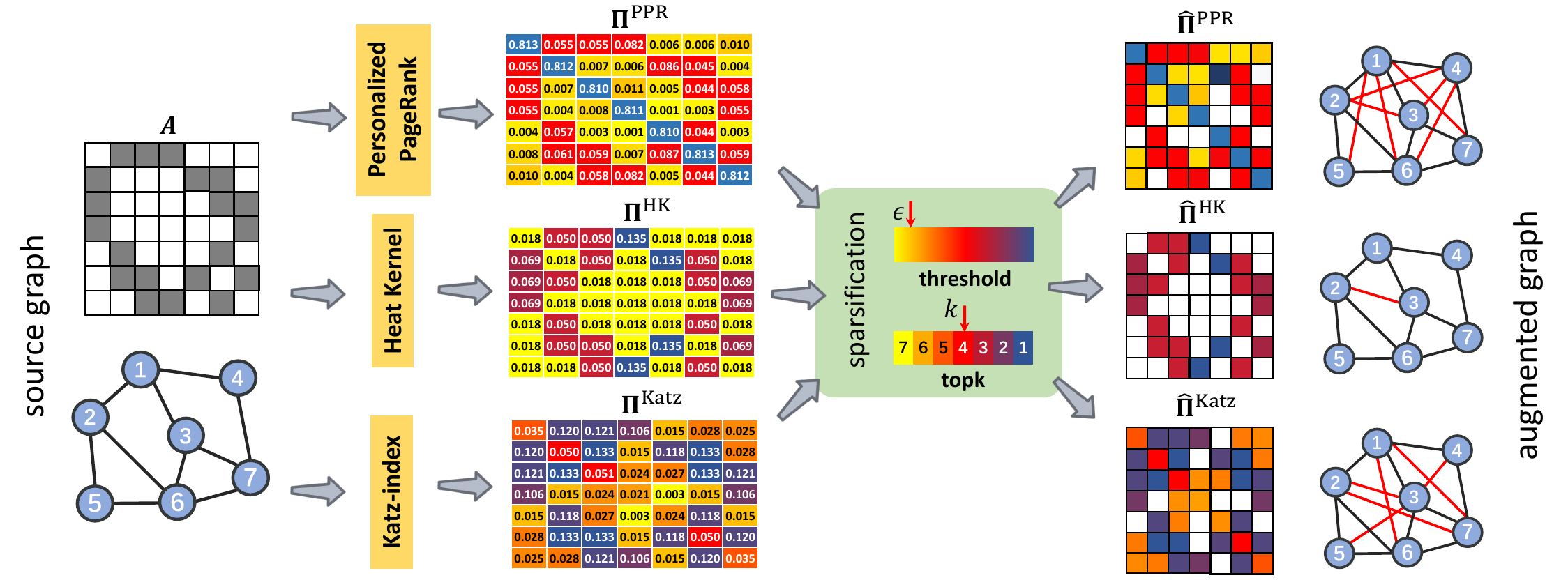}
		\caption{Illustration of different graph propagation augmentations.}
		\label{fig: graph propagation}
\end{figure*}

%
{The technique simulates the diffusion process for propagating node information throughout the graph, effectively capturing long-range dependencies between nodes and smoothing local irregularities. By incorporating broader contextual information, it enriches node representations, thereby facilitating the model's acquisition of higher-order graph information.}
\begin{mydef}{~{Graph Propagation}}{DEFexample}
    For a given graph $G=\left(\boldsymbol{A,X}\right)$, graph propagation measures the proximity between any two nodes via global probabilistic transition, thereby injecting high-order topological information into graph adjacency, yielding an augmented global view $\hat{G}=(\hat{\boldsymbol{\Pi}},\boldsymbol{X})$, \ie
    \begin{equation}\label{eq: graph-propagation}
        \begin{array}{l}
            \boldsymbol{\Pi} =\sum_{k=0}^{\infty} \theta_\textit{k} \cdot \boldsymbol{T}^\textit{k} \cdot \boldsymbol{P}^\top \vspace{3pt}\\
            \hat{\boldsymbol{\Pi}} = \textsf{Sparsification}{(\boldsymbol{\Pi})}
        \end{array}
    \end{equation}
    where $\boldsymbol{\Pi}$ is the generated propagation matrix, $\textsf{Sparsification}(\cdot)$ is a sparsification function, $\theta_\textit{k}$ is the weighting coefficient that controls the ratio of global-local information, $\boldsymbol{T} \in \mathbb{R}^{|\mathcal{V}| \times |\mathcal{V}|}$ is the transition matrix, $\boldsymbol{P}\in \mathbb{R}^{|\mathcal{V}| \times |\mathcal{V}|}$ is formed by stacking the teleport location probability distribution vectors of all nodes and satisfies $\left\lVert \boldsymbol{P}_\textit{i}\right\rVert_2 = 1$ for all $v_\textit{i}\in \mathcal{V}$.
\end{mydef}
Here we summarize several graph propagation instantiations specified by different settings, as listed in Table~\ref{tb: graph-propagation-strategy} and illustrated in Fig.~\ref{fig: graph propagation}.
An earlier study has shown that employing higher-order message propagation mechanisms can significantly improve the performance of graph learning~\cite{GDC-2019}, which inspired the use of generalized graph diffusion methods such as \textbf{Personalized PageRank (PPR)}~\cite{pagerank-1999} and \textbf{Heat Kernel (HK)}~\cite{heatkernel-2007} to generate higher-order augmented views~\cite{MVGRL-2020,MV-GCN-2021,SelfGNN-2021,MERIT-2021}.
The graph diffusion powered by PPR is defined by setting $\boldsymbol{T}=\boldsymbol{AD}^{-1}$ and $\theta_\textit{k}=\alpha(1-\alpha)^\textit{k}$, and HK corresponds to choosing $\boldsymbol{T}=\boldsymbol{AD}^{-1}$ and $\theta_\textit{k}=e^\textit{-t} \cdot \frac{t^\textit{k}}{k!}$, where $\alpha \in (0,1)$ is the tunable teleport probability in random walk, and $t$ denotes the diffusion time.
Notably, compared to the teleport operation with equal probability in PageRank ($\boldsymbol{P} = \frac{1}{|\mathcal{V}|} \cdot \boldsymbol{1}$), PPR is special in that each node has a user-defined (personalized) teleport location probability distribution $\boldsymbol{P}_\textit{i}$.
In addition, SelfGNN~\cite{SelfGNN-2021} also uses the \textbf{Katz-index}~\cite{katz-1953} to capture high-order topological information and generates augmented views.
The Katz-index characterizes the relative importance of nodes from a global perspective by weighting and integrating the reachable paths of different lengths between nodes.
We unify it into the generalized graph propagation formula by setting $\boldsymbol{T} = \boldsymbol{A}$ and $\theta_\textit{k} =  \beta^\textit{k}$, where $\beta$ is the attenuation factor that controls the path weights.

Finally, it is worth noting that the graph propagation augmentation will yield a dense propagation matrix $\boldsymbol{\Pi}$~\cite{GDC-2019}, in which the elements represent the influence between all pairs of nodes. 
To guarantee the sparsity of the final augmented graph, there are two tricks in practice: 1) threshold-$\epsilon$, which sets elements below $\epsilon$ to zero; 2) top-$k$, which keeps the $k$ elements with the largest values per column.
Both of the two sparsification tricks help to truncate small values in $\boldsymbol{\Pi}$, yielding a sparse propagation matrix $\hat{\boldsymbol{\Pi}}$ that can provide a global view during contrastive learning.
\begin{table*}[htp]
    \renewcommand\arraystretch{1.3}
    \centering
    \caption{Summary of different graph propagation augmentations.}
    \label{tb: graph-propagation-strategy}
    \resizebox{\textwidth}{!}{%
    \begin{tabular}{ccccl}
    \hline\hline
    \multirow{2}{*}{\textbf{Reference}}                 & \multicolumn{3}{c}{\textbf{Parameter Setting}}                                                                                 & \multicolumn{1}{c}{\multirow{2}{*}{\textbf{Propagation Equation}}}  \\
    \cline{2-4}
                                               & \multicolumn{1}{c}{$\theta_\textit{k}$}        & \multicolumn{1}{c}{$\boldsymbol{T}$} & \multicolumn{1}{c}{$\boldsymbol{P}$} &                             \\
    \hline
    PageRank~\cite{pagerank-1999}              & ~~~~$\alpha(1-\alpha)^\textit{k}$~~~~          & ~~~~$\boldsymbol{AD}^{-1}$~~~~               & $\frac{1}{|\mathcal{V}|} \cdot \boldsymbol{1}$   &  $\boldsymbol{\Pi} =\sum_{k=0}^{\infty} \alpha(1-\alpha)^\textit{k} \cdot \left(\boldsymbol{AD}^{-1}\right)^\textit{k} \cdot \frac{1}{|\mathcal{V}|} \cdot \boldsymbol{1}  = \alpha(\boldsymbol{I}-(1-\alpha) \boldsymbol{AD}^{-1})^{-1} \cdot \frac{1}{|\mathcal{V}|} \cdot \boldsymbol{1}$                                               \\
    Personalized PageRank~\cite{pagerank-1999} & ~~~~$\alpha(1-\alpha)^\textit{k}$~~~~          & ~~~~$\boldsymbol{AD}^{-1}$~~~~               & personalized $\boldsymbol{P}_\textit{i}$         &  $\boldsymbol{\Pi} =\sum_{k=0}^{\infty} \alpha(1-\alpha)^\textit{k} \cdot \left(\boldsymbol{AD}^{-1}\right)^\textit{k} \cdot \boldsymbol{P}^\top = \alpha(\boldsymbol{I}-(1-\alpha) \boldsymbol{AD}^{-1})^{-1} \cdot \boldsymbol{P}^\top$                                                \\
    Heat Kernel~\cite{heatkernel-2007}         & $e^\textit{-t} \cdot \frac{t^\textit{k}}{k !}$ & ~~~~$\boldsymbol{AD}^{-1}$~~~~               & ~~~~identity matrix $\boldsymbol{I}$~~~~         &  $\boldsymbol{\Pi} =\sum_{k=0}^{\infty} e^\textit{-t} \cdot \frac{t^\textit{k}}{k !} \cdot \left(\boldsymbol{AD}^{-1}\right)^\textit{k} = \exp \left(t \boldsymbol{A D}^{-1}-t\right)$                                               \\
    Katz-index~\cite{katz-1953}                & $\beta^\textit{k}$                             & $\boldsymbol{A}$                             & ~~~~identity matrix $\boldsymbol{I}$~~~~         &  $\boldsymbol{\Pi} =\sum_{k=1}^{\infty} \beta^\textit{k} \boldsymbol{A}^\textit{k} = (\boldsymbol{I}-\beta \boldsymbol{A})^{-1} \cdot \beta \boldsymbol{A}$                                              \\
    \hline\hline
    \end{tabular}}
\end{table*}

\subsubsection{{\textbf{Graph Coarsening} and \textbf{Refinement}}}
{Graph coarsening is a traditional graph reduction technique that hierarchically merges nodes and edges to reduce the size of a graph~\cite{hashemi2024comprehensive}. 
Initially, this method was designed to summarize large graphs, offering top-down views to facilitate visualization and understanding~\cite{tian2008ksnap}. 
Analogous to CV, these hierarchical views of a graph can be likened to multi-resolution representations of an image. 
In addition, graph coarsening is usually followed by a reverse process called graph refinement, which assigns coarse-grained node features to fine-grained nodes.
Therefore, we present the general definition of graph coarsening and refinement as follows.
}
\begin{mydef}{~{Graph Coarsening and Refinement}}{DEFexample}
    {For a graph $G = (\boldsymbol{A}, \boldsymbol{X})$, graph coarsening merges and groups nodes and their associated edges in $G$ to coarse-grained supernodes with reconnected edges through an assignment matrix $\boldsymbol{C}$, yielding a coarse graph $\hat{G}=(\hat{\boldsymbol{A}}, \hat{\boldsymbol{X}}_\textit{coarse})$ with reduced size, \ie}
    \begin{equation}
        \hat{\boldsymbol{X}}_\textit{coarse}={\textsf{rowNorm}(\boldsymbol{C}^\top)}\cdot{\boldsymbol{X}}, \quad\quad   \hat{\boldsymbol{A}}={\boldsymbol{C}}^\top{\boldsymbol{A}}{\boldsymbol{C}}
    \end{equation}
    where $\boldsymbol{C}=\{0,1\}^{|\mathcal{V}|\times |\hat{\mathcal{V}}|}$ and $\boldsymbol{C}_\textit{ij}=1$ indicates that node $v_\textit{i}$ from $G$ is assigned to node $v_\textit{j}$ in the coarse graph $\hat{G}$, $\textsf{rowNorm}(\cdot)$ is applied to perform row-wise max-min normalization. 
    And for the coarse graph $\hat{G}$, graph refinement assigns features of supernodes to their corresponding fine-grained nodes in the original graph $G$, \ie:
    \begin{equation}
        \begin{array}{c}
            {\hat{\boldsymbol{X}}_\textit{refine}=\boldsymbol{C}\hat{\boldsymbol{X}}_\textit{coarse} }
        \end{array}
    \end{equation}
\end{mydef}
\begin{figure*}[htp]
	\centering
		\includegraphics[width=0.9\textwidth]{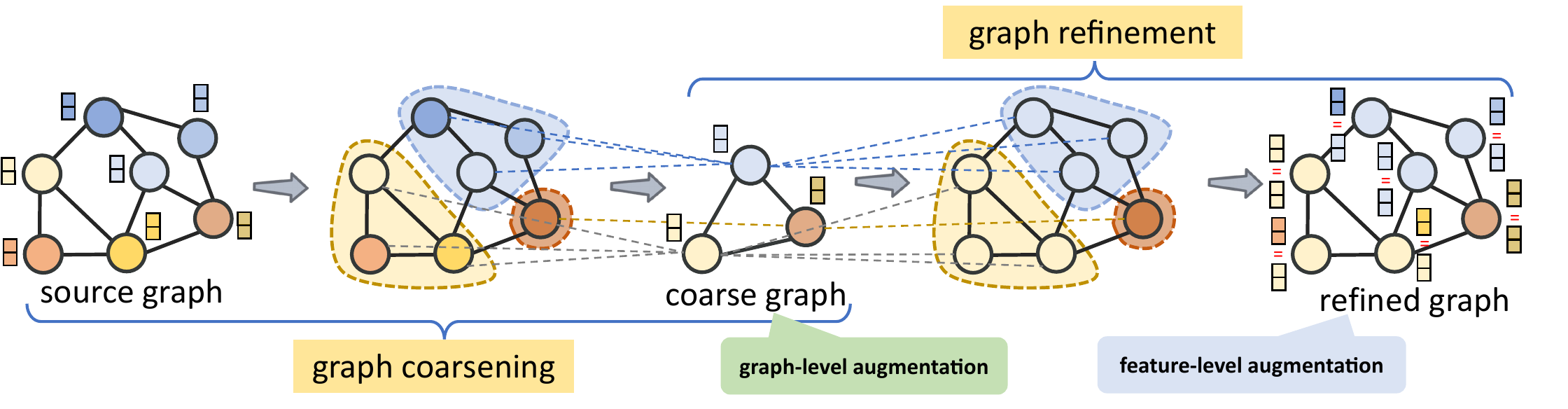}
		\caption{{Illustration of graph coarsening augmentations.}}
		\label{fig: subgraph-sampling}
\end{figure*}

{Note that graph coarsening can be regarded as graph-level augmentation, while subsequent graph refinement can be regarded as feature-level augmentation. The latter can reassign the higher-order coarse features back to the nodes in the original graph, thereby augmenting their feature space.
There are two existing studies on graph coarsening related to data augmentation.
HARP~\cite{chen2018harp} sequentially coarsens a graph through multiple levels. At each level, HARP initializes node representation by refinement and adopts graph embedding methods to attain structure information in current level. This approach, which employs various graphs at each iteration, effectively augments the representations of training data. MILE~\cite{liang2021mile} enhances this process by replacing the embedding methods used in HARP with GNNs and designing a structure-based loss function to improve the quality and efficiency of embedding in each level.}

\subsubsection{\textbf{Graph Mixup}}
This technique aims to generate synthetic graphs by mixing the existing graphs via linear interpolation. Compared with node mixup, graph mixup can be more intuitively analogous to Mixup~\cite{Mixup} in CV. Existing studies~\cite{Mixup,Manifold-Mixup,How-mixup} have demonstrated that Mixup can work well on regular, well-aligned and Euclidean data such as images. 
{However, the success of mixup in image augmentation cannot be easily reproduced in the graph domain due to several key challenges: 
1) The irregular nature of graph structural data makes node alignment difficult; 
2) The topological diversity of different graphs may result in the mixed graph losing important features; 
3) The complexity of graph connectivity and message-passing mechanisms requires the simultaneous mixing of node features and topological information, avoiding interference with the feature aggregation process of other nodes. Recent studies~\cite{2branch-mixup,IGM-2024,OMG-2023,ifmixup-2023,S-Mixup-g-2023,GAMT-2022,G-Mixup-2022,FGWMixup-2024} have made various attempts to enhance the adaptability and effectiveness of mixup techniques in graph learning.}
Here, we first provide a general definition of graph mixup:
\begin{mydef}{~{Graph Mixup}}{DEFexample}
    {For a pair of graphs $G_\textit{i} = (\boldsymbol{A}_\textit{i}, \boldsymbol{X}_\textit{i}, Y_\textit{i})$ and $G_\textit{j} = (\boldsymbol{A}_\textit{j}, \boldsymbol{X}_\textit{j}, Y_\textit{j})$, graph mixup first aligns them in augmentation space $\mathbb{G}$ and then performs linear interpolation to generate an augmented graph $\hat{G}$ with label $\hat{Y}$, \ie}
    \begin{equation}
        \begin{array}{c}
            {(\boldsymbol{G}_\textit{i}, \boldsymbol{G}_\textit{j}) = \mathsf{Alignment} (G_\textit{i}, G_\textit{j})} \vspace{3pt}\\
            {\hat{G} \sim P\left(\hat{G} \mid (1-\lambda) \cdot \boldsymbol{G}_\textit{i} + \lambda \cdot \boldsymbol{G}_\textit{j} \right)} \vspace{3pt}\\
            {\hat{Y} = (1-\lambda) \cdot Y_\textit{i} + \lambda \cdot Y_\textit{j}}
        \end{array}
    \end{equation}
    {where $\mathsf{Alignment}(\cdot,\cdot) $ is the alignment operation and $\boldsymbol{G}_\textit{i},\boldsymbol{G}_\textit{j}\in \mathbb{G}$.}
\end{mydef}

\begin{table*}[htp]
    \renewcommand\arraystretch{1.2}
    \centering
    \caption{{Summary of different graph mixup augmentations.}}
    \label{tb: graph-mixup}
    \resizebox{\textwidth}{!}{%
    \begin{tabular}{ccc} 
    \hline\hline
    \textbf{{Reference}}         & \textbf{{Alignment Mechanism}}                                  & \textbf{{Augmentation Space}}  \\ 
    \hline
    {Two-branch Mixup~\cite{2branch-mixup}, IGM~\cite{IGM-2024}, OMG~\cite{OMG-2023}} & {map graphs to the same embedding space via GNN and Pooling   } & {embedding           }         \\
    {ifMixup~\cite{ifmixup-2023}                                                    } & {introduce virtual nodes to align the size of graphs          } & {input               }         \\
    {S-Mixup-g~\cite{S-Mixup-g-2023}                                                } & {generate a soft assignment matrix for aligning two graphs    } & {input               }         \\
    {GAMT~\cite{GAMT-2022}                                                          } & {apply graph coarsening to align the size of graphs           } & {embedding, attention}         \\
    {G-Mixup~\cite{G-Mixup-2022}                                                    } & {estimate the graphons and interpolate in the graphon space   } & {graphon             }         \\
    {FGWMixup~\cite{FGWMixup-2024}                                                  } & {optimization in the distance space defined by the FGW metrics} & {metric              }         \\
    \hline\hline
    \end{tabular}}
\end{table*}
{Note that \emph{the key to graph mixup augmentation lies in designing effective alignment mechanisms that make linear interpolation on graphs feasible and rational}, as illustrated in Table~\ref{tb: graph-mixup}.
For example, ifMixup~\cite{ifmixup-2023} first aligns two graphs of different sizes by introducing virtual nodes with all-zero feature vectors, and then generates a new graph by mixing the two graphs in topological space.
Two-branch Mixup~\cite{2branch-mixup}, IGM~\cite{IGM-2024}, and OMG~\cite{OMG-2023} utilize graph encoders and pooling operations to map a pair of graphs into the same embedding space and perform interpolation on the graph representations for augmentation. 
G-Mixup~\cite{G-Mixup-2022} first estimates graphons for different classes of graphs and then performs interpolation in the graphon space to obtain mixed graphons, finally augmenting synthetic graphs by sampling from the mixed graphons. 
S-Mixup-g~\cite{S-Mixup-g-2023} first computes a soft assignment matrix between two graphs via a graph matching network, then uses this matrix to align one of the graphs to the other, and finally mixes them to generate the augmented graph. 
GAMT~\cite{GAMT-2022} coarsens all input graphs to the same size and performs interpolation operations during the calculation of attention and value in the graph transformer to achieve data augmentation. 
FGWMixup~\cite{FGWMixup-2024} searches for the optimal match between the synthetic and source graphs in the distance space defined by the Fused Gromov-Wasserstein metric, ensuring that the synthetic graphs effectively preserve the features and structural information of the source graphs.}

%% file: 4-6-label.tex
\subsection{Label-level Augmentation}\label{sec: label-aug}
Most of \gda~ strategies mainly manipulate the features and structure of existing graphs to achieve augmentation, without specific constraints on labels.
While label-level augmentation is another important techniques used to augment the limited labeled data using the unlabeled data.
Without loss of generality, we present a general definition of label augmentation.

\begin{mydef}{~{Label Augmentation}}{DEFexample}
    {For a partially labeled dataset $\mathcal{D}=\mathcal{D}_\textit{l} \cup \mathcal{D}_\textit{u}$, where $\mathcal{D}_\textit{l}$ is the labeled set and $\mathcal{D}_\textit{u}$ is the unlabeled set, the general workflow of label augmentation proceeds as follows:
    \begin{enumerate}[leftmargin=18pt]
        \item \textbf{Model Initialization}: pre-train models as pseudo-label predictors using $\mathcal{D}$;
        \item \textbf{Pseudo-Label Generation}: use the pre-trained models to predict labels for the unlabeled data $\mathcal{D}_\textit{u}$, yielding a pseudo-labeled set $\mathcal{D}_\textit{p}$;
        \item \textbf{Data Selection}: select a subset $\mathcal{D}_\textit{p}^\textit{s}$ from the pseudo-labeled set $\mathcal{D}_\textit{p}$ via certain criteria;
        \item \textbf{Data Augmentation}: combine the selected pseudo-labeled set $\mathcal{D}_\textit{p}^\textit{s}$ with the original labeled set $\mathcal{D}_\textit{l}$ to form an augmented labeled set $\mathcal{D}_\textit{l}^\textit{aug} = \mathcal{D}_\textit{l} \cup \mathcal{D}_\textit{p}^\textit{s}$.
    \end{enumerate}}
\end{mydef}
{After label augmentation, the augmented set $\mathcal{D}_\textit{l}^\textit{aug}$ can be used to retrain the models. And the process of label augmentation and retraining can go through multiple iterations, where the models are continuously updated as new pseudo-labeled data is generated and added.}

\begin{table*}[htb]
    \renewcommand\arraystretch{1.3}
    \centering
    \caption{Summary of different label-level augmentations.}
    \label{tb: pseudo-labeling-strategy}
    \resizebox{\textwidth}{!}{%
    \begin{tabular}{ccccc} 
    \hline\hline
    \textbf{Type (custom)}        & \textbf{Reference}                     & \textbf{Model Initialization (predictor)}                 & \textbf{Pseudo-Label Generation}                                       & \textbf{Data Selection}              \\ 
    \hline
              \multirow{3}{*}{threshold}   & MEvolve     & graph classification model                       & same as the original graph                                    & \multirow{3}{*}{threshold}  \\ 
    \cdashline{2-4}
               & AutoGRL                                  & label propagation algorithm (LPA)                & label propagation                                             &                             \\ 
    \cdashline{2-4}
               & NRGNN                                  & GNN for node classification~                     & semi-supervised prediction via GNN                            &                             \\ 
    \hdashline
                 \multirow{2}{*}{clustering} & CGCN & GNNs for node classification and node clustering & same as the highest conﬁdence labeled node in its cluster     & \multirow{2}{*}{match}      \\ 
    \cdashline{2-4}
                 & M3S                                  & k-means algorithm, GNN for node classification   & same as the label of the closest cluster of a certain class   &                             \\ 
    \hdashline
      \multirow{2}{*}{sharpen} & GraphMix, GRAND   & GNN for node classification                      & sharpen the average predictions across multiple augmentations & \multirow{2}{*}{/}          \\ 
    \cdashline{2-4}
                & NASA                                  & GNN for node classification                      & sharpen the average of its neighbors’ predictions~            &                             \\
    \hline\hline
    \end{tabular}}
\end{table*}
\begin{figure*}[htp]
	\centering
		\includegraphics[width=\textwidth]{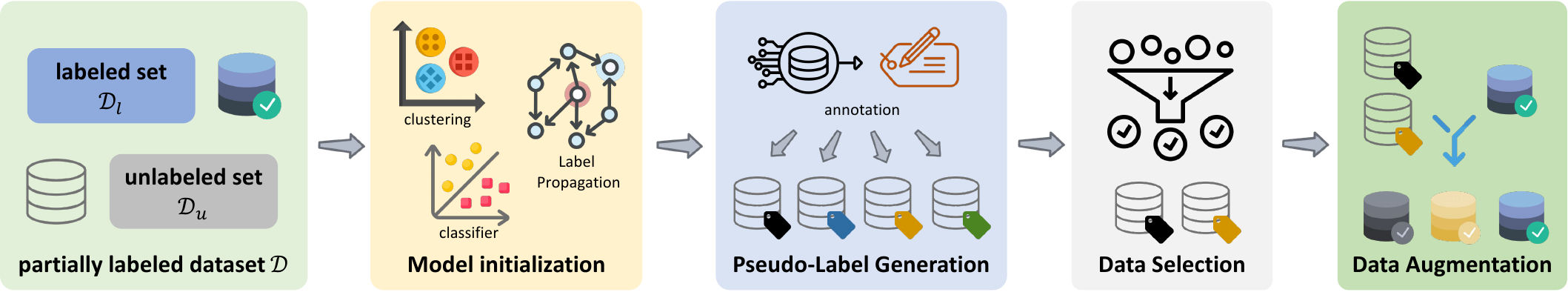}
		\caption{{The general workflow of label augmentation proceeds as follows: 
        1) pre-train predictors using partially labeled dataset;
        2) predict pseudo labels for unlabeled data using predictors;
        3) select partial pseudo-labeled samples via certain criteria;
        4) combine selected pseudo-labeled and labeled data to form an augmented set.}}
		\label{fig: pseudo-labeling}
\end{figure*}

We summarize several graph related works using label augmentation and divide them into three sub-categories, as listed in Table~\ref{tb: pseudo-labeling-strategy} and illustrated in Fig.~\ref{fig: pseudo-labeling}.
The \textbf{threshold-based} methods mainly construct pseudo-labels via model predictions and determine whether a pseudo-labeled sample has high confidence by comparing the model prediction probability with a threshold.
M-Evolve~\cite{MEvolve-tnse} first generates the virtual graphs via edge rewiring and assigns them the labels of the original graphs as pseudo-labels. Then a concept of label reliability is introduced based on the intuition that pseudo-labels generally have higher reliability when matched to predictions. M-Evolve finally uses the label reliability threshold to filter out the virtual graphs with high label reliability as the augmented data.
NRGNN~\cite{NRGNN-2021} first inserts the missing edges between labeled and unlabeled nodes through edge prediction, then trains a GNN model on the rewired graph and uses the predictions of unlabeled nodes as their pseudo-labels, and finally retains the unlabeled nodes with their pseudo-labels whose predicted probability is greater than a threshold as augmented data.
The \textbf{clustering-based} methods mainly use cluster assignments as pseudo-labels by introducing unsupervised clustering tasks, and then filter pseudo-labeled samples with high confidence by matching the consistency of supervised predictions and cluster assignments.
M3S~\cite{M3S-2020} first runs K-means clustering on the node embeddings, then aligns labeled and unlabeled clusters by comparing the distance of centroids between clusters, and finally passes the label of the labeled cluster to the nearest unlabeled cluster.
CGCN~\cite{CGCN-2020} first learns clusters by GMM-VGAE model, then select the highest confidence (softmax prediction score) labeled sample of each class, and finally passes their labels to the unlabeled nodes in the clustering network.

The above two techniques are obsessed with generating high-confidence pseudo-labels for unlabeled data, which usually fail when unlabeled data has low-confidence predictions. While sharpen-based label augmentation does not rely on high-confidence predictions and is suitable for more unlabeled scenarios.
It is commonly used in conjunction with consistency regularization~\cite{NASA-2022,Graphmix-2021,GRAND-2020}.
GraphMix~\cite{Graphmix-2021} applies the average prediction on multiple random perturbations of an input unlabeled sample along with sharpening trick~\cite{DL-2015}, further augments training data and improve prediction accuracy.
GRAND~\cite{GRAND-2020} utilizes sharpening trick to construct labels for unlabeled nodes based on the average prediction over multiple data augmentation, further achieving consistency regularization.
NASA~\cite{NASA-2022} proposes a neighbor-constrained regularization to enforce the predictions of neighbors to be consistent with each other, in which the sharpening trick is used to generate label for the center node based on the average predictions of its neighbors.

\subsection{{Summary}}
{After discussing the GDAug techniques at various scales, we summarize them in this subsection, highlighting their multifaceted differences, as shown in Table~\ref{tb: compare}.}

\begin{table*}[htp]
    \renewcommand\arraystretch{1.3}
    \centering
    \caption{{Comparison of different graph data augmentation techniques.}}
    \label{tb: compare}
    \resizebox{\textwidth}{!}{%
    \begin{tabular}{ccccc} 
    \hline\hline
    \textbf{Scale}          & \textbf{GDAug}        & \textbf{Prior Knowledge}                                    & \begin{tabular}[c]{@{}c@{}}\textbf{Manipulated }\\\textbf{Objects}\end{tabular} & \textbf{Application}                 \\ 
    \hline
    \multirow{5}{*}{Micro}  & Feature Shuffling     & Variability of node features                                & $\boldsymbol{X}$                                                                               & node-level                     \\ 
                            & Feature Masking       & Uncertainty in node features                                & $\boldsymbol{X}$                                                                               & node-level, graph-level  \\ 
                            & Node Dropping         & Redundancy or low information content of nodes              & $\mathcal{V}$                                                                            & graph-level                    \\ 
                            & Node Mixup            & Smoothness of node features and labels                      & $\boldsymbol{X}, y$                                                                            & node-level                     \\ 
                            & Edge Rewiring         & Noisy, dense or sparse graph structure                      & $\mathcal{E}$                                                                            & node-level, graph-level  \\ 
    \hline
    \multirow{2}{*}{Meso}   & Subgraph Sampling     & Key local structures                                        & $\boldsymbol{A}, \boldsymbol{X}$                                                                           & graph-level                    \\ 
                            & Subgraph Substitution & Replace subgraphs that are similar in structure or function & $\boldsymbol{A}, \boldsymbol{X}$                                                                            & graph-level                    \\ 
    \hline
    \multirow{4}{*}{Macro}  & Graph Propagation     & Long distance dependence and smoothness of information flow &    $\boldsymbol{A}$                                                                             & node-level                     \\ 
                            & Graph Coarsening      & Preserve global information of the graph                    & $\boldsymbol{A}, \boldsymbol{X}$                                                                            & node-level, graph-level  \\ 
                            & Graph Mixup           & Dependent on alignment mechanism                            & $\boldsymbol{A}, \boldsymbol{X}, Y$                                                                         & graph-level                    \\ 
                            & Label Augmentation    & Labels with high predictive confidence are usually valuable & $y$ or $Y$                                                                            & node-level, graph-level  \\
    \hline\hline
    \end{tabular}}
\end{table*}

%% file: 5-domain.tex
\section{Domain-specific Complex Graph Data Augmentation}

{Section ~\ref{sec: gda} summarizes the \gda~ methods, which are primarily proposed for simple graphs. However, in real-world scenarios, to adequately describe the complex interaction characteristics among entities in specific domains, the graph construction process often includes additional elements, such as entity and relationship types, entity descriptions, interaction time, and spatial locations, ultimately forming different types of complex graphs, such as heterogeneous graphs, temporal graphs, spatio-temporal graphs, and hypergraph graphs. Therefore, this section will integrate complex graph types and application domains to summarize the relevant \gda~ methods.}

\subsection{{Heterogeneous Graph Data Augmentation}}

{Heterogeneous graphs encompass multiple types of nodes and edges, effectively simulating the complex systems and relationships in the real world. 
They are particularly suited for processing and integrating information from different domains, facilitating connections between various types of information through a multitude of relationships.
This affords a rich context for data analysis and knowledge extraction, rendering them extensively employed in fields such as recommendation systems, knowledge graphs, and social network analysis.
}

{Due to their inherent complexity, heterogeneous graphs often encounter challenges associated with low-quality data, which tends to be more severe compared to that observed in homogeneous graphs. 
Several studies~\cite{HRGCN-2023,GAAD-2024,MuHca-2023} directly apply existing \gda~ techniques to heterogeneous graphs. 
For example, HRGCN~\cite{HRGCN-2023} utilizes random and heuristic \textbf{edge rewiring} techniques to augment anomalous heterogeneous data, thereby enhancing the generalizability of heterogeneous graph anomaly detection models. 
GAAD~\cite{GAAD-2024} quantifies the edge existence between heterogeneous nodes using a graph encoding-decoding model and strengthens the graph structure by adding highly possible edges.
MuHca~\cite{MuHca-2023} employs \textbf{node mixup} techniques to generate hard negative samples for contrastive learning in heterogeneous graphs. 
However, these existing data augmentation techniques are primarily designed for simple homogeneous graphs and seldom consider the heterogeneity of graphs.
Applying these techniques directly to heterogeneous graphs might disrupt the high-order semantic information, thereby affecting the effectiveness of subsequent graph learning.
Recently, several studies~\cite{HeCo-2021,MAHGA-2023,MEOW-2024,MCL-2023,MVSE-2021,A2-CLM-2024} have utilized \textbf{meta-paths} or \textbf{meta-graphs}, which can describe the high-order semantics of heterogeneous graphs, integrating them with existing \gda~ techniques to explore effective data augmentation for heterogeneous graphs.
For instance, methods like HeCo~\cite{HeCo-2021}, MEOW~\cite{MEOW-2024}, and MCL~\cite{MCL-2023} initially construct contrastive views based on meta-paths or meta-path induced subgraphs, and then employ data augmentation strategies such as \textbf{feature masking} and \textbf{edge dropping} to introduce noise and enhance data diversity. 
MAHGA~\cite{MAHGA-2023} first constructs meta-path reachable graphs according to different types of meta-paths respectively, then extracts graphons from these graphs, and finally performs \textbf{interpolation} on different graphons to augment meta-paths. 
A2-CLM~\cite{A2-CLM-2024} performs \textbf{subgraph sampling} to obtain meta-graphs containing specific semantics from software samples, using these as challenging augmentation instances to enhance malware detection.}

\subsection{{Temporal Graph Data Augmentation}}


{Temporal graphs are capable of describing the temporal interactions between entities by capturing the emergence, evolution, and disappearance of entities and their interactions over time, thereby revealing temporal behavioral patterns in complex systems. Owing to their time sensitivity and high complexity, temporal graphs have been widely applied in fields such as social network analysis, financial market monitoring, and bioinformatics. 
}



{However, in real-world scenarios, temporal graphs often contain substantial temporal noise due to systemic delays and errors during data collection. For example, in social platforms, email communications between users might be delayed due to server lag. In blockchain finance, the timestamp of a transaction tends to reflect the time when the transaction was packaged into a block, which may differ from the actual initiation time of the transaction. To reduce the impact of temporal noise on the graph learning process, existing studies~\cite{MeTA-2021,TF-GCL-2022,TGAC-2023,TGEditor-2023,TACL-2023,TGCL4SR-2024} typically combine temporal information with existing \gda~ techniques to design data augmentation strategies for temporal graphs.
MeTA~\cite{MeTA-2021} employs \textbf{feature masking} techniques to add constrained Gaussian noise to timestamps on edges, simulating the common time shift in real scenarios without altering the order of interactions.
TGCL4SR~\cite{TGCL4SR-2024} first alleviates data sparsity through \textbf{subgraph sampling}, and then adopts a similar approach as MeTA by adding Gaussian noise to the timestamp attribute of edges to simulate temporal noise.
TF-GCL~\cite{TF-GCL-2022} avoids the decay of interaction influence over time by randomly sampling a portion of interactions and replacing timestamps with the current time.
TGAC~\cite{TGAC-2023} quantifies the importance of edges by combining node centrality and the time of edge occurrence, and generates augmented views by removing edges of low importance.}

\subsection{{Spatio-Temporal Graph Data Augmentation}}


{Spatio-temporal graphs enable the description and analysis of temporal changes and interactions among entities across both spatial and temporal dimensions in the real world. Comprising nodes representing different entities or locations, edges representing connections or interactions between entities, and timestamps, these graphs capture not only static connection patterns but also embed information about the temporal evolution of relationships. Consequently, they are extensively employed in analyzing traffic road systems and urban regions.}
%


{However, spatio-temporal graphs constructed from real scenarios often suffer from data noise, incompleteness and spatial heterogeneity due to issues such as sensor malfunctions and spatially independent zoning, which are not favorable for downstream spatio-temporal mining applications. Existing studies~\cite{STDGI-2019,AutoST-2023,GraphST-2023,ST-SSL-2023,USTGCL-2024,STGCL-2022} usually design data augmentation strategies from both temporal and spatial dimensions to alleviate the low-quality problem.
STGCL~\cite{STGCL-2022} improves data quality by \textbf{interpolating} data across continuous time steps to capture intermediate state data, and by transforming time-series input into the frequency domain followed by second-order neighborhood smoothing to reduce high-frequency noise. 
AutoST~\cite{AutoST-2023} designs a denoising variational graph autoencoder to reconstruct the structure of spatio-temporal graphs, enabling automatic learning of inter-regional dependencies. 
ST-SSL~\cite{ST-SSL-2023} employs \textbf{feature masking} to randomly remove low-correlation traffic volumes against temporal noise, and designs inter-regional heterogeneity metric to guide inter-regional topological reconnection against spatial noise. 
USTGCL~\cite{USTGCL-2024} simulates temporal noises such as sensor malfunctions through threshold-based \textbf{feature masking} and enhances local and global correlations between regions through \textbf{edge rewiring}.}

\subsection{{Hypergraph Data Augmentation}}



{Unlike simple graphs, hypergraphs can represent higher-order information by defining hyperedges that connect more than two nodes. Such feature enables hypergraphs to capture complex multi-dimensional relationships, offering richer expressive capabilities compared to simple graphs. As a result, hypergraphs are widely used in scenarios such as social networks and recommendation systems for mining higher-order group relationships.}


{To address the low-quality data problem, existing hypergraph studies~\cite{HHGR-2021,HyperGCL-2022,HCCF-2022,TriCL-2023} typically incorporate existing \gda~ techniques to optimize the representation of higher-order information.
HHGR~\cite{HHGR-2021} designs a double-scale \textbf{node dropping} strategy on user-level hypergraph to create self-supervision signals that can regularize user representations with different granularities against the sparsity issue.
HyperGCL~\cite{HyperGCL-2022} designs a hypergraph generation model to parameterize hypergraph augmentation based on \textbf{edge rewiring}, which aids hypergraph contrastive learning in automatically generating effective augmented views to enrich self-supervision signals.
HCCF~\cite{HCCF-2022} alleviates the overfitting problem in hypergraph contrastive collaborative filtering by applying \textbf{edge dropping} to the user-item interaction graph and the corresponding hypergraph. 
TriCL~\cite{TriCL-2023} utilizes four data augmentation strategies in hypergraph contrastive learning, including \textbf{node dropping}, \textbf{hyperedge removing}, \textbf{membership dropping} (dropping nodes from hyperedges), and node \textbf{feature masking}.}

%% file: 6-metric.tex
\section{Evaluation Metrics and Design Guidelines}\label{sec: metrics}
In this section, we summarize existing metrics and guidelines for evaluating or designing \gda~ techniques.
We first introduce several evaluation metrics that are commonly used in graph analysis tasks, and then introduce some new metrics and guidelines specifically designed for \gda.

\subsection{Common Metrics}

Since \gda~ is an auxiliary technique, existing studies often measures its effectiveness by evaluating the performance improvements in downstream tasks achieved with its assistance.
For classification tasks like node classification~\cite{MeTA-2021,NodeAug-2020} and graph classification~\cite{MEvolve-cikm,Ethident-2022}, it is common to use accuracy-based metrics such as \textbf{Accuracy}, \textbf{Precision}, \textbf{F1 score}, \textbf{Area Under Curve (AUC)} and \textbf{Average Precision (AP)} for reflecting the classification performance from different perspectives.
For graph representation learning~\cite{InfoGraph-2020,GCA-2021,SUBG-CON,MVGRL-2020}, the learned embeddings will be fed into the downstream classifiers or clustering algorithms, and evaluated via accuracy-based metrics as mentioned above, or clustering-based metrics such as \textbf{Normalized Mutual Information (MNI)} and \textbf{Adjusted Rand Index (ARI)}.
Additionally, several works apply \gda~ in pre-training task~\cite{GCC-2020} or recommendation~\cite{SGL-2021}, followed by evaluation with ranking-based metrics like \textbf{HITS@$k$}.

\subsection{Specific Metrics and Design Guidelines}
In addition to the aforementioned common metric, several special metrics and guidelines have also been proposed for \gda.
Here we present the detailed definitions and descriptions of them.


\subsubsection{\textbf{Change Ratio}}
The metric quantifies the degree of modifications to the graph structure or features resulting from augmentation techniques such as node dropping~\cite{MH-Aug-2021}, edge rewiring~\cite{MEvolve-tnse,MH-Aug-2021} and feature masking~\cite{GRACE-2020}. It can also be regarded as a probabilistic parameter or manipulation budget to guide the augmentation process. Formally, we have the general definition of change ratio:
\begin{mydef}{~{Change Ratio}}{DEFexample}
    Given a graph $G=(\mathcal{V}, \mathcal{E}, \boldsymbol{X})$ and its augmentation $\hat{G}=(\hat{\mathcal{V}}, \hat{\mathcal{E}}, \hat{\boldsymbol{X}})$, change ratio quantifies the proportion of modifications made to the fundamental elements of the source graph, \ie
    \begin{equation}
        \begin{array}{l}
            \mathcal{M}_{\Delta \mathcal{V}} = (|\mathcal{V} - \hat{\mathcal{V}}| + |\hat{\mathcal{V}} - \mathcal{V}|) / |\mathcal{V}| \vspace{3pt}\\
            \mathcal{M}_{\Delta \mathcal{E}} = (|\mathcal{E} - \hat{\mathcal{E}}| + |\hat{\mathcal{E}} - \mathcal{E}|) / |\mathcal{E}, \quad\quad \mathcal{M}_{\Delta \boldsymbol{X}} = \Vert \hat{\boldsymbol{x}} - \boldsymbol{x}\Vert_0 / F
        \end{array}
    \end{equation}
   {where $F$ is the dimension of feature vector, and $\Vert \cdot\Vert_0$ is the zero norm.}
\end{mydef}

\subsubsection{\textbf{Tradeoff between Consistency and Diversity}}
NASA~\cite{NASA-2022} proposes consistency and diversity metrics to measure the correctness and generalization ability of \gda, respectively. 
\begin{mydef}{~{Consistency vs. Diversity}}{DEFexample}
    For two models $f_\theta$ and $\hat{f}_\theta$ trained by the training set $\mathcal{D}_\textit{train}$ and the augmented set $\hat{\mathcal{D}}_\textit{train}$, respectively, the consistency represents the accuracy of $\hat{f}_\theta$ on the validation set $\mathcal{D}_\textit{val}$, \ie
    \begin{equation}
        \mathcal{M}_\textit{c} = \operatorname{Acc}\left(\hat{f}_\theta \left(\mathcal{D}_\textit{val}\right), \mathcal{Y}_\textit{val}\right)
    \end{equation}
    where $\mathcal{Y}_\textit{val}$ is the labels of validation set.
    The diversity represents the prediction difference between $f_\theta$ and $\hat{f}_\theta$ on the validation set, \ie
    \begin{equation}
        \mathcal{M}_\textit{d} = \Vert \hat{f}_\theta \left(\mathcal{D}_\textit{val}\right) - f_\theta \left(\mathcal{D}_\textit{val}\right) \Vert^2_\textit{F}
    \end{equation}
    where $\Vert \cdot\Vert_\textit{F}$ is the Frobenius norm.
\end{mydef}
Specifically, a lower consistency indicates that the augmentation hurts the original data distribution, but a higher consistency may not contribute well to the generalization of the model while maintaining correctness.
On the other hand, a lower diversity indicates that the augmentation contributes little to the generalization of the model, while a higher diversity cannot ensure the correctness of the augmentation.
Neither metric alone can fully evaluate the quality of augmentation. 
Therefore, NASA combines these two metrics to trade off the design of \gda, expecting to achieve augmentation with better correctness and generalization, improve the performance of augmented models, and build generalized decision boundaries.

\subsubsection{\textbf{Tradeoff between Affinity and Diversity}}
\citet{trivedi2022augmentations} state that data augmentation should generate samples that are sufficiently close to the source data to share task-relevant semantics, while maintaining enough diversity to prevent information redundancy. Therefore, they utilize the affinity and diversity metrics defined in~\cite{gontijo2020tradeoffs} to balance data augmentation.
\begin{mydef}{~{Affinity vs. Diversity}}{DEFexample}
    For a model $f_\theta$ trained from $\mathcal{D}_\textit{train}$, the affinity can be measured by the ratio of the accuracy on the augmented validation set $\hat{\mathcal{D}}_\textit{val}$ to the accuracy on the source validation set $\mathcal{D}_\textit{val}$, \ie
    \begin{equation}
        \mathcal{M}_\textit{A} = \operatorname{Acc}\left(f_\theta (\hat{\mathcal{D}}_\textit{val}), \mathcal{Y}_\textit{val}\right)/\operatorname{Acc}\left(f_\theta (\mathcal{D}_\textit{val}), \mathcal{Y}_\textit{val}\right)
    \end{equation}
    where $\mathcal{Y}_\textit{val}$ is the labels of validation set.
    And the diversity can be measured by the ratio of the final training loss on the augmented training set $\hat{\mathcal{D}}_\textit{train}$, relative to the final training loss on the source training set $\mathcal{D}_\textit{train}$, \ie
    \begin{equation}
        M_\textit{D} = \mathbb{E}[\hat{\mathcal{L}}_\textit{train}] / \mathbb{E}[\mathcal{L}_\textit{train}]
    \end{equation}
\end{mydef}
Specifically, the affinity metric is used to quantify the distribution shift of the augmented data compared to the source data, with a lower affinity indicating that the augmented data is out-of-distribution for the model.
On the other hand, the diversity metric quantifies how difficult it is for a model to learn from augmented data rather than the source data.

\subsubsection{\textbf{Preserving Connectivity}}
Several studies~\cite{MEvolve-tnse,SubMix-2022} state that the connectivity information of a graph before and after augmentation should not be changed, as defined below.
\begin{mydef}{~{Preserving Connectivity}}{DEFexample}
    For a graph $G$ and its augmentation $\hat{G}$, $\hat{G}$ should follow the connectivity information of $G$, \ie $\hat{G}$ should be connected if and only if $G$ is connected.
\end{mydef}

%% file: 7-application.tex
\section{Applications of Graph Data Augmentation}\label{sec: application}
In this section, we review and discuss how \gda~ improves graph learning from two application levels, \ie data and model.

\subsection{Data-level Application}
Collecting and constructing graph-structured data on real systems inevitably suffers from several dilemmas, such as label scarcity, class imbalance, information redundancy, \gsb{noise}, \etc, which directly lead to low-quality graph data and indirectly lead to poor graph learning performance. 
\gda~ technology has been proposed to alleviate the problems of over-fitting, weak generalization, and low fairness caused by low-quality graph data at the data level.

\subsubsection{Label Scarcity}
In practical scenarios, obtaining data labels requires human labor and is time-consuming and laborious, leading to the label scarcity problem.
For example, labeling account types in financial transaction networks is subject to privacy restrictions due to sensitive identity information;
Toxicity labeling of molecular graphs requires extensive toxicology detection experiments;
Labeling documents in citation networks requires summarizing their topics in terms of their content.
Graph learning methods tend to fall into over-fitting and weak generalization when working on small and sparsely labeled graph datasets.
To alleviate the issue, data augmentation is a prevalent remedy that can expand data distribution and increase data diversity, achieving an improvement in the generalization power of machine learning models trained on augmented data.

Among the aforementioned \gda~ techniques, label-level augmentation combined with \textbf{graph self-training} works well as a general solution to improve semi-supervised graph learning when training data is limited.
Specifically, graph self-training can generate high-confidence pseudo labels for unlabeled data as supervision via pre-trained models trained with limited labeled data, and the augmented data with pseudo labels can be used to retrain pre-trained models or train new models.
Representative works include M3S~\cite{M3S-2020}, CGCN~\cite{CGCN-2020}, NRGNN~\cite{NRGNN-2021} and M-Evolve~\cite{MEvolve-tnse}, as discussed in Sec.~\ref{sec: label-aug}.
Moreover, \gda~ has also been applied in \textbf{graph self-supervised learning (GSSL)}~\cite{Survey-GSSL-1}.
For contrastive GSSL, \gda~ is generally used to generate augmented views for each instance. Two views generated from the same instance are generally regarded as a positive pair, while those generated from different instances are generally regarded as a negative pair.
For example, GraphCL~\cite{GraphCL-2020} proposes four \gda~ strategies, including feature masking, node dropping, edge rewiring and subgraph sampling, to generate augmented views for graphs. 
Other representative works include GCA~\cite{GCA-2021}, MVGRL~\cite{MVGRL-2020}, MERIT~\cite{MERIT-2021}, CSSL~\cite{CSSL-2021}, GBT~\cite{GBT-2022}, \etc~
For generative GSSL, it first uses \gda~ to mask partial features or structures of graph data, then uses pretext tasks such as reconstruction to take the masked features or structures as self-supervised signals.
For example, ~\citet{hu2020strategies} first mask node and edge attributes, and then use the pretext task of attribute prediction to capture the domain knowledge of molecular graphs.

\subsubsection{Class Imbalance}
Class imbalance is another form of label scarcity when there is an unequal distribution of classes in the training data.
In other words, the labeled minority classes may have significantly fewer samples than the majority classes.
This problem is extremely common in practice and can be observed in various research fields such as anomaly detection and fraud detection~\cite{survey-class-imbalance}.
For example, in the financial transaction network, most accounts belong to normal users, while the number of abnormal or fraudulent accounts labeled is far less than normal accounts.
Since most existing graph learning methods are mainly based on the class-balance assumption, directly training graph models on the class-imbalanced data cannot learn the features of the minority class samples well, resulting in sub-optimal performance and low fairness.
To alleviate the issue in graph data, \gda~ can be used to balance the class distribution.
Existing works mainly utilize node interpolation augmentation to generate synthetic nodes for minority classes, such as GraphMixup~\cite{GraphMixup-2022}, GraphENS~\cite{GraphENS-2021}, and GraphSMOTE~\cite{Graphsmote-2021}, as described in Sec.~\ref{sec: node-interpolation}.
\subsubsection{Structural Noise}
Real-world graphs generally contain noisy and task-irrelevant edges, which will interfere with message propagation and aggregation in graph learning, resulting in sub-optimal performance.
For example, the automatic following of bot accounts in social networks affects the characterization of user preferences by graph algorithms; the key structures that determine a certain property of a molecule often only occupy a small part of the entire molecular graph.
In this regard, edge-level augmentations are used in \textbf{graph structure learning} to optimize the noisy graph structure and learn more robust graph representation.
For example, Luo~\cite{PTDNet-2021} \etal proposed a learnable topological denoising network to remove task-irrelevant edges, further improving the robustness and generalization of GNNs.
Other representative works include NeuralSparse~\cite{NeuralSparse-2020}, TO-GNN~\cite{TO-GNN-2019}, RobustECD~\cite{RobustECD-2021}.

\subsubsection{Generalizability}

Existing GRL methods usually train dedicated models for domain-specific data and have weak transferability to out-of-distribution (OOD) data.
In this regard, several existing works utilize subgraph sampling augmentations for \textbf{graph pre-training} and \textbf{transfer learning}.
For example, GCC~\cite{GCC-2020} performs random-walk sampling to augment the ego-net subgraphs, and EGI~\cite{EGI-2021} utilizes ego-net sampling and information maximization for training transferable GNNs.

\subsection{Model-level Application}
Despite the excellent performance of GRL methods in characterizing the features of graph data, they still expose many weaknesses and limitations, such as over-smoothing on deep GNNs, vulnerability to graph adversarial attacks, and poor transferability.
In this regard, integrating \gda~ techniques with existing graph learning paradigms can partially alleviate these model limitations.

\subsubsection{Over-smoothing}
When the over-smoothing problem arises, node representations gradually become indistinguishable as network depth increases, eventually losing relevance to the input features and resulting in vanishing gradients. 
Several studies~\cite{DropEdge-2020,Deepgcns,li2018deeper,Hou2020Measuring} have investigated the issue of over-smoothing in GNNs. Some of them have employed \gda~ techniques to perturb either the graph topology or features, effectively alleviating the over-smoothing problem. 
For instance, DropEdge~\cite{DropEdge-2020} introduces edge removal augmentation to randomly disrupt the message propagation process. 
AdaEdge~\cite{chen2020measuring} utilizes adaptive edge rewiring augmentation to iteratively optimize the graph topology by removing inter-class edges and adding intra-class edges based on model predictions. 
GRAND~\cite{GRAND-2020} introduces random perturbations in the message propagation process via node feature dropping, thereby augmenting the node's receptive field during representation learning.

\subsubsection{Vulnerability}
GNNs have been demonstrated to inherit the vulnerability of deep neural networks~\cite{GAL-survey}, meaning they are susceptible to manipulation by small input perturbations referred to as adversarial attacks. To tackle this issue, some studies integrate \textbf{graph adversarial learning} and \gda~ to acquire robust graph representations.
For instance, FLAG~\cite{FLAG-2022} employs gradient-based feature masking during training to iteratively optimize node features, thereby ensuring graph models remain invariant to minor input perturbations and further enhancing their robustness and generalization.
GROC~\cite{GROC-2021} employs gradient-based edge rewiring as an adversarial transformation during graph contrastive learning to generate augmented views, thereby enhancing the robustness of GNNs against adversarial attacks. 
GraphCL~\cite{GraphCL-2020} conducts adversarial experiments to demonstrate that graph contrastive learning with \gda~ can effectively enhance the robustness of GNNs against multiple evasion attacks.


%% file: 8-challenge-future.tex
\section{Open Issues and Future Directions}\label{sec: challenges}
Despite the considerable attention and widespread application of \gda~ techniques, there remain several shortcomings and challenges in current research. This section summarizes open issues and discusses future research directions.

\subsection{{Complex Graph Data Augmentation}}
{Current \gda~ techniques are primarily designed for simple graphs. Although some techniques exist for complex graphs, they often reuse methods developed for simple graphs without modifications. Such practice overlooks the unique characteristics and requirements of complex graphs, thereby limiting the effectiveness and applicability of \gda~ techniques. For instance, the heterogeneity, multi-relational nature, and dynamics in complex graphs require more customized augmentation strategies. Future research should focus on developing \gda~ techniques specifically tailored for complex graphs, considering their unique structures and attributes to enhance the quality of augmented data and model performance. Additionally, systematic evaluation frameworks should include assessments of applicability to complex graphs to ensure the effectiveness of augmentation techniques across various complex scenarios. This will drive significant advancements in \gda~ technologies, thereby expanding their impact across a broader spectrum of applications.}

\subsection{{Interpretable Graph Data Augmentation}}
{Current \gda~ methods often rely on certain prior knowledge, lacking in-depth interpretability during the design and application phases. Many augmentation techniques are proposed without theoretical justification for their design rationale, which limits the understanding of their effectiveness and applicability in practical applications. This lack of interpretability not only affects the credibility of the methods but also restricts their adoption and application across different domains and tasks.
Additionally, there is a significant concern regarding the preservation of graph semantics during augmentation. Many existing methods may compromise the inherent semantics of the graph when modifying its structure, such as randomly deleting or adding nodes, thereby undermining the meaningful properties of the original graph.
In future research, there should be a greater emphasis on the interpretability of \gda, explicitly elucidating the underlying design principles and mechanisms of each technique. Simultaneously, methods that ensure semantic preservation during augmentation should be developed to maintain the integrity and relevance of the graph's original information. Enhancing interpretability through theoretical analysis and experimental validation, along with focusing on semantic consistency, will facilitate the development of more transparent, credible, and semantically robust augmentation methods, improving their effectiveness and applicability in real-world applications.}

\subsection{{Scalable Graph Data Augmentation}}
{Research on scalable \gda~ is crucial but has received limited attention in current studies. The primary challenge lies in efficiently applying augmentation techniques to large-scale graphs containing millions or even billions of nodes and edges. Most existing methods are computationally intensive and may not scale well, limiting their practical application in handling large datasets. This scalability issue hinders the widespread adoption and effectiveness of \gda~ techniques in fields such as social network analysis, bioinformatics, and recommendation systems. Future research should focus on developing scalable \gda~ techniques that maintain computational efficiency without sacrificing the quality of augmented data. This could involve leveraging advanced parallel processing, distributed computing, and efficient sampling techniques. Additionally, it will be crucial to develop scalable algorithms that can dynamically adapt augmentation strategies based on the size and complexity of graphs.}

\subsection{{Comprehensive Evaluation System}}
{In current research, evaluating the quality and effectiveness of \gda~ remains a formidable challenge. Most existing studies rely on general performance metrics to measure the effectiveness of \gda, but these methods usually assess augmentation quality indirectly through downstream task performance rather than directly evaluating the augmented samples. Although some studies~\cite{NASA-2022,trivedi2022augmentations,gontijo2020tradeoffs} have proposed consistency and diversity metrics to assess the correctness and generalizability of \gda, these metrics are often a combination of predictive indicators with limited interpretability. The lack of a comprehensive evaluation system makes it difficult to accurately assess the specific impact of augmented data on model performance. Additionally, the absence of standardized metrics and benchmarks hinders direct comparison of different augmentation techniques and their contributions to model robustness and generalization. Therefore, future research should focus on developing systematic evaluation frameworks to comprehensively assess the impact of \gda~ techniques on data and models, ensuring the generation of high-quality augmented data. This is crucial for advancing \gda~ technology and facilitating its effective application across various tasks and datasets.}



%% file: 9-conclusion.tex
\section{Conclusion}
In this paper, we present a comprehensive survey of graph data augmentation (\gda).
Specifically, we classify \gda~ methods into six categories according to multi-scale graph elements, i.e., feature-level, node-level, edge-level, subgraph-level, graph-level, and label-level augmentations. We then summarize several common performance metrics and specific design metrics for evaluating \gda.
Furthermore, we review and discuss the data-level and model-level applications of \gda.
Finally, we outline existing open issues as well as future directions in this field.